\def\year{2022}\relax
\documentclass[letterpaper]{article} 
\usepackage{aaai22}  
\usepackage{times}  
\usepackage{helvet}  
\usepackage{courier}  
\usepackage[hyphens]{url}  
\usepackage{graphicx} 
\usepackage{natbib}  
\usepackage{appendix}
\usepackage{caption} 
\DeclareCaptionStyle{ruled}{labelfont=normalfont,labelsep=colon,strut=off} 
\usepackage{algorithm}
\usepackage{algorithmic}

\usepackage{booktabs}
\usepackage{amssymb}
\usepackage{amsmath}
\usepackage{xr-hyper}
\usepackage{hyperref}

\usepackage{cleveref}
\usepackage{lipsum}
\newtheorem{theorem}{Theorem}
\newtheorem{lemma}{Lemma}
\crefname{lemma}{Lemma}{Lemmas}

\usepackage{graphics}

\renewcommand{\cite}[1]{\citep{#1}}

\makeatletter
\newcommand*{\addFileDependency}[1]{
  \typeout{(#1)}
  \@addtofilelist{#1}
  \IfFileExists{#1}{}{\typeout{No file #1.}}
}
\makeatother

\newcommand*{\myexternaldocument}[1]{%
    \externaldocument{#1}%
    \addFileDependency{#1.tex}%
    \addFileDependency{#1.aux}%
}
\myexternaldocument{supplementary_material}
\usepackage{newfloat}
\usepackage{listings}
\floatstyle{ruled}
\newfloat{listing}{tb}{lst}{}
\floatname{listing}{Listing}
\title{KerGNNs: Interpretable Graph Neural Networks with Graph Kernels}

\author {
    Aosong Feng\textsuperscript{\rm 1},
    Chenyu You\textsuperscript{\rm 1},
    Shiqiang Wang\textsuperscript{\rm 2},
    Leandros Tassiulas\textsuperscript{\rm 1}
}
\affiliations {
    \textsuperscript{\rm 1} Yale University, New Haven, CT, USA\\
    \textsuperscript{\rm 2}  IBM T. J. Watson Research Center, Yorktown Heights, NY, USA\\
    aosong.feng@yale.edu, chenyu.you@yale.edu, wangshiq@us.ibm.com,
    leandros.tassiulas@yale.edu
}

\begin{document}

\maketitle
\begin{abstract}
Graph kernels are historically the most widely-used technique for graph classification tasks. However, these methods suffer from limited performance because of the hand-crafted combinatorial features of graphs. In recent years, graph neural networks (GNNs) have become the state-of-the-art method in downstream graph-related tasks due to their superior performance. Most GNNs are based on Message Passing Neural Network (MPNN) frameworks. However, recent studies show that MPNNs can not exceed the power of the Weisfeiler-Lehman (WL) algorithm in graph isomorphism test. To address the limitations of existing graph kernel and GNN methods, in this paper, we propose a novel GNN framework, termed \textit{Kernel Graph Neural Networks} (KerGNNs), which integrates graph kernels into the message passing process of GNNs. Inspired by convolution filters in convolutional neural networks (CNNs), KerGNNs adopt trainable hidden graphs as graph filters which are combined with subgraphs to update node embeddings using graph kernels. In addition, we show that MPNNs can be viewed as special cases of KerGNNs. We apply KerGNNs to multiple graph-related tasks and use cross-validation to make fair comparisons with benchmarks. We show that our method achieves competitive performance compared with existing state-of-the-art methods, demonstrating the potential to increase the representation ability of GNNs. We also show that the trained graph filters in KerGNNs can reveal the local graph structures of the dataset, which significantly improves the model interpretability compared with conventional GNN models\footnote{https://github.com/asFeng/kergnns}.
\end{abstract}

\noindent 
\begin{figure}[t]
  \centering \includegraphics[width=0.98\linewidth]{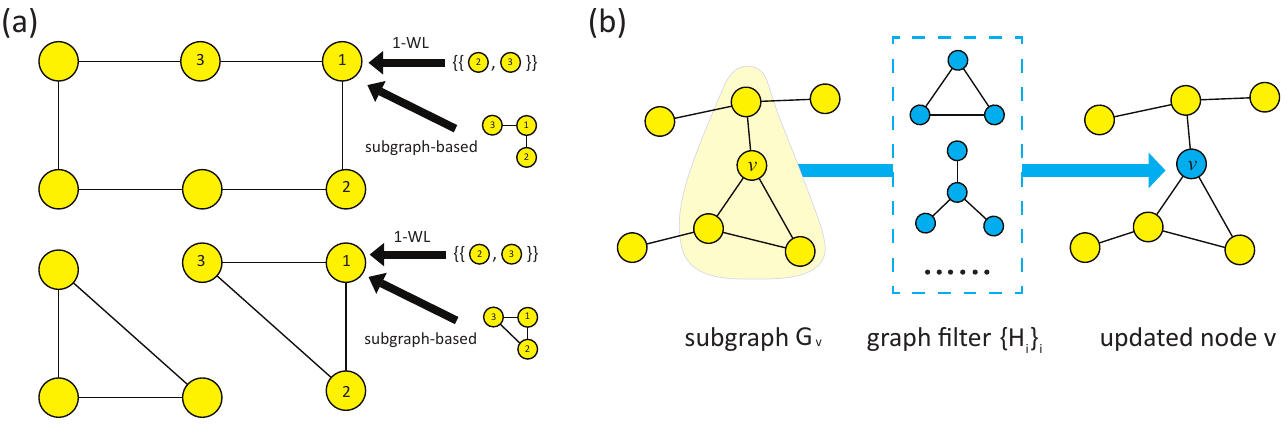}
  \caption{(a) 1-WL graph isomorphism test cannot distinguish one hexagon and two triangles because of same neighborhood multisets, while subgraph-based method can find the difference based on different subgraph topologies. (b) The yellow shadow represents the subgraph of node $v$. After interacting with graph filters, the updated node is colored in blue.}
  \label{fig:concept}
  \vspace{-0.3in}
\end{figure}

In recent years, the machine learning research community has devoted substantial energy to applying graph neural networks (GNNs) to numerous downstream graph-related tasks
\cite{ying2018hierarchical,kipf2016semi,zhang2018link,chen2017supervised}.
Considering graph-structured tasks, the commonalities between different variants of GNNs are Message Passing Neural Networks (MPNNs) \cite{gilmer2017neural}. MPNNs consist of two stages, including neighborhood aggregation and graph-level readout. Specifically, for neighborhood aggregation, there are three steps for each node to generate embeddings: (1) receiving messages from its neighbors, (2) aggregating messages, and (3) updating its own features to encode the local structural information.
For graph-level tasks, a permutation-invariant readout function is used to extract feature representations from the entire graph. 
In fact, MPNNs have been motivated and derived as a continuous and differentiable analog of the Weisfeiler-Lehman (WL) algorithm \cite{leman1968reduction} which is known to successfully test graph isomorphism for a broad class of graphs. However, recent studies \cite{xu2018powerful,morris2019weisfeiler} show that MPNNs are at most as powerful as the WL kernel \cite{shervashidze2011weisfeiler} and WL algorithm regarding the graph isomorphism tests. This demonstrates theoretical limits in the expressivity of popular GNNs. For example, Figure \ref{fig:concept}(a) shows two graphs which cannot be distinguished by 1-WL algorithm, and therefore are also indistinguishable by MPNNs.

Before the advent of GNNs, graph kernels were the most widely-used techniques for solving graph classification tasks \cite{kriege2020survey}. Graph kernels measure the similarity between graphs, and can be applied into a kernel machine (e.g., support vector machine). Kernel functions remove the need of learning node embeddings in high dimensions, and enable us to operate in a high dimensional feature space by simply computing the kernel value in the low-dimensional feature space, which is more computationally efficient than computing in the high-dimensional space directly. Because of the empirical success of kernel-based methods and the increasing availability of graph-structured datasets, numerous graph kernel methods have been proposed, including walks and paths kernels \cite{gartner2003graph,kashima2003marginalized,borgwardt2005shortest}, subgraph kernels \cite{shervashidze2009efficient}, and WL kernels \cite{shervashidze2011weisfeiler}. However, graph kernels still have limitations due to their hand-crafted features and fixed feature construction scheme, which may not effectively capture high-dimensional information (e.g., complex node interactions) on large graphs.

In this paper, to address the above-mentioned issues and increase the expressivity of GNNs, we propose a \textbf{subgraph-based node aggregation algorithm} by combining GNNs and graph kernels into one framework, and thus the advantages of both methods can be leveraged. On one hand, for neighborhood aggregation, we apply graph kernels which use the subgraph induced by node neighbors, so that the expressivity will not be limited by 1-WL isomorphism test which uses the multiset of neighboring nodes. An example is shown in Figure \ref{fig:concept}(a), where we note that node 1 in both graphs has the same neighborhood multiset but induce different subgraph topologies with its neighbors which can be distinguished by graph kernels. On the other hand, we make the feature construction scheme of graph kernels trainable following the standard GNN training framework, possibly allowing for greater adaptability. 

Based on the subgraph-based node aggregation, we propose a novel GNN framework, termed \emph{KerGNNs}. Specifically, we first introduce a set of trainable hidden graphs, named graph filters, in each layer. Each node within the input graph is associated with a subgraph capturing its local topological information. We then adopt graph kernel functions to compare the similarity of graph filters and input subgraphs, and use the computed kernel values to update the respective node's feature representations (as shown in Figure~\ref{fig:concept}(b)).
We show that KerGNNs provide a new kernel perspective to extend the standard CNN structure into the graph domain and generalize most MPNNs.  
The proposed model is then evaluated with various real-world graph and node classification tasks, and the results show superior performance of KerGNNs compared with many existing state-of-the-art models. To better understand the predictions of GNN-based methods, KerGNNs can further visualize the trained graph filters, similar to visualizing filters in CNNs, and thus provide better human-interpretable explanations for
a variety of graph-related tasks, compared to existing GNNs. \textbf{Our main contributions} are summarized as follows: 

\begin{enumerate}
\item We use neighborhood subgraph topology combined with kernel methods for GNN neighborhood aggregation, and show with proof that the expressivity of this approach is not limited by the 1-WL algorithm.
\item We provide a new perspective to generalize CNNs into the graph domain, by showing that both 2-D convolution and graph neighborhood aggregation can be interpreted using the language of kernel methods.
\item Besides envisioning the output graphs of the model, KerGNNs can further reveal the local structure of the input graphs by visualizing the topology of trained graph filters, which significantly improves the model interpretability and transparency compared with standard MPNNs. 

\end{enumerate}

\section{Related Work}
\subsection{Expressivity}
Several works have been devoted to improving the expressivity of GNNs by introducing spatial, hierarchical, and higher-order GNN variants. For example, \citet{abu2019mixhop} proposed the mix-hop structure which can learn a more general class of neighborhood mixing relationships. \citet{sato2019approximation} proposed to use Consistent Port Numbering GNN to augment the neighborhood aggregation, but port orderings are not unique and different orderings may lead to different expressivity. \citet{klicpera2020directional} leveraged the atom coordinate information in the molecular graph to improve the expressivity, but the notion of direction is hard to generalize to more general graphs. 
\citet{nguyen2020graph} used the graph homomorphism numbers as updated embeddings and show the expressivity of such graph classifiers with universality property, which unfortunately lacks neural network structure.
Higher-order GNN variants have been studied in \citet{morris2019weisfeiler} and \citet{maron2019provably}, which is more powerful than the 1-WL graph isomorphism test. However, higher-order methods always involve heavy computation and KerGNNs introduce a different way to break this 1-WL limit.

\subsection{Combination of Graph Kernel and GNNs} Graph kernels and GNNs can be combined in the same framework. Some works apply graph kernels and neural networks at different stages \cite{navarin2018pre,nikolentzos2018kernel}. There are also works on using GNN architecture to design new kernels. For example, \citet{du2019graph} proposed a graph kernel equivalent to infinitely wide GNNs which can be trained using gradient descent.

A different line of research focuses on integrating kernel methods into GNNs.
\citet{lei2017deriving} mapped inputs to RKHS by comparing inputs with reference objects. However, the reference objects they use lack graph structure and may not be able to capture the structural information.
\citet{chen2020convolutional} proposed GCKN which maps the input into a subspace of RKHS using walk and path kernel. While GCKN utilizes the local walk and path only starting from the central node, our model considers any walks (up to a maximal length) within the subgraph around the central node, and can thus explore more topological structures.
Another recent work by \citet{rwgnn} focused on improving model transparency by calculating the graph kernels between trainable hidden graphs and the entire graph. However, the method only supports a single-layer model and lacks theoretical interpretation. Our KerGNN model generalizes their scenario by applying hidden graphs to extract local structural information instead of the entire graph, and therefore constructs a multi-layer structure with better graph classification performance.

\subsection{Explainability} 
Both graph structures and feature information lead to complex GNN models, making it hard for a human-intelligible explanation of the prediction results. Therefore, the transparency and explainability of GNN models are important issues to address. \citet{baldassarre2019explainability} compared two main classes of explainability methods using infection and solubility problems. \citet{pope2019explainability} introduced explainability methods for the popular graph convolutional neural networks and demonstrated the extended methods on visual scene graphs and molecular graphs. \citet{ying2019gnnexplainer} proposed a model-agnostic approach that can identify a compact subgraph that has a crucial role in GNN's prediction. In addition to visualizing output graphs as in regular GNNs, our KerGNN provides trained hidden graphs as a byproduct of training without additional computations, which contain useful structural information showing the common characteristics of the whole dataset instead of one specific graph, and can be helpful for interpreting the predictions of GNNs.

\section{Background: Graph Kernels}

\label{sec:graphkernel}
Graph kernels have been proposed to solve the problem of assessing the similarity between graphs, and therefore making it possible to perform classification and regression with graph-structured data. Most graph kernels can be written as the sum of several pair-wise base kernels, following the ${\mathcal{R}}$-convolution framework \cite{haussler1999convolution}:
\begin{equation}
\label{eq:rtype}
\begin{split}
     K(G,G')=\sum_{v\in\mathcal{V}}\sum_{v'\in\mathcal{V'}}k_{\mathrm{base}}(v,v'),
\end{split}
\end{equation}
where $G=(\mathcal{V},\mathcal{E})$, $G'=(\mathcal{V'},\mathcal{E'})$ are two input graphs with node attributes, and $k_{\mathrm{base}}$ can be any positive definite kernel defined on the node attributes. In this paper, we mainly consider random walk kernel which will be integrated into our proposed model in the next section.

Random walk kernels are one of the most studied graph kernels. They count the number of walks that two graphs have in common, and were initially proposed by \citet{gartner2003graph} and \citet{kashima2003marginalized}. Among numerous variations of the random walk kernel, we deploy the $P$-step random walk kernel which compares random walks up to length $P$ in two graphs. 


Following Equation \ref{eq:rtype}, we can write the base kernel of random walks with length $p$ as 
\begin{align*}
&k_{\mathrm{base}}^{p}(v,v') \\
&\!\!=
    \begin{cases} \langle{a(v),a(v')}\rangle, &\!\!\! \text{if } p=0\\
    \langle{a(v),a(v')}\rangle\cdot\lambda\!\!\!\sum\limits_{u\in{\mathcal{N}(v)}}\sum\limits_{u'\in{\mathcal{N}(v')}}\!\!\!\!k_{\mathrm{base}}^{(p-1)}(u,u'),&\!\!\! \text{if } p>0 \end{cases}
\end{align*}
where $\lambda$ is the coefficient, $\mathcal{N}(v)$ denotes neighbors of $v$, $p$ denotes the length of random walks which we compare in two graphs. If $p=0$, the random walk kernel is equivalent to the simple node-pair kernel. To efficiently compute the random walk kernels, we follow the generalized framework of computing walk-based kernel \cite{vishwanathan2006fast}, and utilize the direct product graph defined as below.

\textbf{Definition 1} (Direct Product Graph). \textit{For two labeled graphs $G=(\mathcal{V},\mathcal{E})$ and $G'=(\mathcal{V}',\mathcal{E}')$, the direct product graph is defined as $G_{\times}=G\times G'=(\mathcal{V}_{\times},\mathcal{E}_{\times})$, defined as
$\mathcal{V}_{\times}=\{(v,v'): v \in \mathcal{V}\land v'\in \mathcal{V}'\}$
and $\mathcal{E}_{\times}=\{\{ (v,v'),(u,u')\}: \{v,u\}\in \mathcal{E} \land \{v',u'\} \in \mathcal{E}' \}$.
}

Performing a random walk on the direct product graph $G_{\times}$ is equivalent to performing the simultaneous random walks on graphs $G$ and $G'$. The $P$-step random walk kernel can be calculated as
\begin{equation}
\label{eq:rweq}
\begin{split}
   K(G,G')=\sum_{p=0}^{P}{K_p(G,G')}=\sum_{p=0}^{P} \lambda_p\sum_{i,j=1}^{|\mathcal{V}_{\times}|} \left[ A_{\times}^{p} \right]_{ij},
\end{split}
\end{equation}
where $A_{\times}$ is the adjacency matrix of $G_{\times}$ and $\lambda=(\lambda_0,\lambda_1,...)$ is a sequence of weights. 
It should be noted that the $(i,j)$-th element of $A_{\times}^p$ (i.e., $A_{\times}$ to the power of $p$) represents the number of common walks of length $p$ between the $i$-th and $j$-th node in $G_{\times}$.
 
To generalize the above formula into the continuous and multi-dimensional scenario, we first define the vertex attributes of the direct product graph $G_{\times}$. Given the node attribute matrix $\mathbf{X}\in \mathbb{R}^{n\times d}$ for a graph with $n$ nodes and each node attribute is of dimension $d$, the node attribute matrix $\mathbf{S}$ of the direct product graph $G_{\times}=G_1\times G_2$ is calculated as $\mathbf{S}=\mathbf{X}_1 \mathbf{X}_2^T$, where $\mathbf{X}_1\in \mathbb{R}^{n_1\times d}$ and $\mathbf{X}_2\in \mathbb{R}^{n_2\times d}$ are the node attribute matrices for $G_1$ and $G_2$, respectively, and $\mathbf{S}\in \mathbb{R}^{n_1\times n_2}$. The $(i,j)$-th element of matrix $\mathbf{S}$ encodes the similarity between the $i$th-node of $G_1$ and the $j$-th node of $G_2$. We flatten $\mathbf{S}$ into vector $\mathbf{s}\in \mathbb{R}^{n_1 n_2}$ for ease of notation, and then integrate the encoded pair-wise similarity into Equation \ref{eq:rweq}
\begin{equation}
\label{eq:ssa}
\begin{split}
   K_p(G,G')=\sum_{i,j=1}^{|\mathcal{V}_{\times}|} \mathbf{s}_i \mathbf{s}_j \left[A_{\times}^{p} \right]_{ij} = \mathbf{s}^T A_{\times}^{p} \mathbf{s}.
\end{split}
\end{equation}
Based on this equation, we can calculate the kernel value between two input graphs using the similarity of common walks as the metric. The details of calculating Equation \ref{eq:ssa} are included in Appendix.

In practice, we also consider a slight variation of Equation~\ref{eq:rtype} by adding trainable weights to each base kernel term, and we call it deep random walk kernel:
\begin{equation}
\begin{split}
     K(G,G')=\sum_{v\in\mathcal{V}}\sum_{v'\in\mathcal{V'}}w_{(v,v')}k_{\mathrm{base}}(v,v'),
\end{split}
\end{equation}
where $w_{(v,v')}$ represents the trainable weight assigned to the base kernel.

\section{Proposed Model}
In this section, we first discuss the framework of the proposed KerGNN model.
Then we introduce the concept of subgraph-based neighborhood aggregation, and use it to analyze the expressivity of KerGNNs.
Next, we show that KerGNNs are inspired by CNNs and compare them from the kernel perspective.
Finally we argue that KerGNNs can generalize MPNN architecture and analyze the time complexity.

\subsection{KerGNN Framework}
\label{sec:proposed_approach}
In this subsection, we introduce the KerGNN model which updates each node's embedding according to the subgraph centered at this node instead of the rooted subtree patterns in MPNNs, as shown in Figure \ref{fig:concept}(b). Unless otherwise specified, we refer to the subgraph as the vertex-induced subgraph formed from a node and all its 1-hop neighbors.

We first define the embeddings of nodes and subgraphs, which are mapping functions from graphs to the feature space and from nodes to the feature space.

\textbf{Definition 2} (Feature mapping). \textit{Given a graph $G = (\mathcal{V},\mathcal{E})$, a node feature mapping is a node-wise mapping function $\phi:\mathcal{V}\to\mathbb{R}^d$, which maps every node $v\in\mathcal{V}$ to a point $\phi(v)$ in $\mathbb{R}^d$, and $\phi(v)$ is called the feature map for node $v$. A graph feature mapping is a function $\Phi:\mathcal{G}\to\mathbb{R}^{d'}$, where $\mathcal{G}$ is the set of graphs, and $\Phi(G)$ is called the feature map for graph $G\in\mathcal{G}$.}

For an $L$-layer neural network, we call the input layer the $0$-th layer. At each hidden layer $l$, the input to this layer is an undirected graph $G=(\mathcal{V},\mathcal{E})$, and each node $v\in \mathcal{V}$ has a feature map $\phi_{l-1}(v)\in \mathbb{R}^{d_{l-1}}$. The output of layer $l$ is the same graph $G$, because we do not consider graph pooling here, and each node $v\in \mathcal{V}$ in the output graph has a feature map $\phi_{l}(v)\in \mathbb{R}^{d_{l}}$. For example, $G$ can be a graph in the dataset, and $\phi_0(v)$ is the node attributes with dimension $d_0$. For graphs with discrete node labels, the attributes can be represented as the one-hot encodings of labels and the dimension of attributes corresponds to the total number of classes. For graphs without node labels, we use the degree of the node as the node attribute.

Inspired by the filters in CNN, we define a set of graph filters at each KerGNN layer to extract the local structural information around each node in the input graph (see Figure~\ref{fig:concept}(b)).

\textbf{Definition 3} (Graph filter). \textit{The $i$-th graph filter at layer $l$ is a graph $H^{(l)}_{i}$ with $n^{(l)}_{i}$ nodes. It has a trainable adjacency matrix $A^{(l)}_{i}\in\mathbb{R}^{n^{(l)}_{i}\times n^{(l)}_{i}}$ and node attribute matrix $W^{(l)}_{i}\in\mathbb{R}^{n^{(l)}_{i}{\times}d_{l-1}}$.}

At layer $l$, there are $d_l$ graph filters such that the output dimension is also $d_l$, and each node attribute in the graph filter, represented by each row of $W^{(l)}_{i}$, has the same dimension as the node feature map $\phi_{l-1}(v)$ in the input graph. 

\subsubsection{KerGNN Layer.}

Now we consider a single KerGNN layer. We assume the input is a graph-structured dataset with undirected graph $G=(\mathcal{V},\mathcal{E})$, and each node $v\in\mathcal{V}$ has the attribute $a(v)\in \mathbb{R}^{d_0}$.  Then the input node feature map is $\phi_0(v)=a(v)$.

Each node $v$ in the graph is equipped with a subgraph $G_v=(\mathcal{V}_v,\mathcal{E}_v)$, and feature maps $\{\phi_0(u):{u\in \mathcal{V}_v}\}$ are transformed to $\phi_1(v)$ in a way such that neighbors' local information (topological information and node representations) contained in $G_v$ will be aggregated to the central node $v$. We then rely on the graph filters $\{H^{(1)}_{i}: i=1,...,d_1 \}$ to obtain $\phi_1(v)$. 
Specifically, we calculate $\phi_1(v)$ by projecting subgraph feature map $\Phi_0(G_v)$ into the $i$-th dimension of $\phi_1(v)$ using the kernel function value between graph filter ${H^{(1)}_{i}}$ and subgraph $G_v$, i.e.,
\begin{equation}
\label{eq:1LayerKernel}
\begin{split}
   \phi_{1,i}(v) = K(G_v,H^{(1)}_i),
\end{split}
\end{equation}
where we adopt a random walk kernel as $K(\cdot,\cdot)$, which is introduced in Equation \ref{eq:rweq}. After calculating the kernel value of the subgraph $G_v$ with respect to every graph filter $\{H^{(1)}_{i}:i=1,...,d_1\}$, we obtain every dimension of node $v$'s feature map $\phi_0(v)$, which forms the output of the KerGNN layer.

It should be noted that using graphs $G_v$ and $H^{(1)}_i$ to calculate the kernel value is equivalent to performing inner product of $\phi_1(G_v)$ and $\phi_1(H^{(1)}_i)$ in an implicit high-dimensional space, and using feature map of $G_v$ instead of the multiset of neighboring nodes (as used in MPNNs) improves expressivity, which is analyzed in the next subsection. Besides, if we use the output space $\mathbb{R}^{d_1}$ to approximate the high-dimensional space introduced by the kernel method, the updating rule will correspond to the convolutional kernel network proposed by \citet{mairal2014convolutional}, and we will follow the same idea when we compare KerGNNs with CNNs in the later subsection.

\subsubsection{Multiple-layer Model.} 
Based on the single-layer analysis above, we can construct a multiple-layer KerGNN by stacking KerGNN layers followed by readout layers. Specifically, the input to layer $l$ is the graph $G$ with node feature map $\{\phi_{l-1}(v): v \in G\}$. Layer $l$ is parameterized by $d_l$ graph filters $\{H^{(l)}_{i}:i=1,...,d_l\}$. Each graph filter $H^{(l)}_{i}$ has a trainable adjacency matrix $A^{(l)}_{i}$ and node attributes $W^{(l)}_{i}$. Then the $i$-th dimension of the output feature map for node $v$ in $G$ can be explicitly calculated as 
\begin{equation}
\begin{split}
   \phi_{l,i}(v) = K(G_v,H^{(l)}_{i}).
\end{split}
\end{equation}
The forward pass of the $l$th-layer of KerGNNs is summarized in Algorithm \ref{alg:kergnn}. 

For the graph classification, we then deploy the graph-level readout layer to generate the embedding for the entire graph. We obtain the graph representation at each layer by summing all the nodes' representations. To leverage information from every layer of the model, we then concatenate the graph representations across all layers:
\begin{equation}
\begin{split}
   \Phi(G) = \mathrm{concat}\left(\sum\nolimits_{v\in G}\phi_{l}(v) \,\Big\vert\, l=0,1,...,L \right).
\end{split}
\end{equation}

\begin{algorithm}[tb]
\small
   \caption{Forward pass in $l$-th KerGNN layer}
   \label{alg:kergnn}
\begin{algorithmic}
   \STATE {\bfseries Input:} Graph $G=(\mathcal{V},\mathcal{E})$; Input node feature maps 
   $\{\phi_{l-1}(v): v\in\mathcal{V}\}$; Graph filters $\{H^{(l)}_{i}: i=1,...,d_l\}$; Graph kernel function $K$ \vspace{0.1in}
   
   \STATE {\bfseries Output:} Graph $G=(\mathcal{V},\mathcal{E})$; Output node feature maps
   $\{\phi_{l}(v): {v\in\mathcal{V}}\}$\vspace{0.1in}
   
   \FOR{$v\in\mathcal{V}$ }
   \STATE $G_v=\mathrm{subgraph}(\{v\}\cup\mathcal{N}(v))$;
   \FOR{$i=1$ {\bfseries to} $d_l$}
   \STATE $\phi_{l,i}(v)=K(G_v,H^{(l)}_{i})$;
   \ENDFOR
   \ENDFOR
\end{algorithmic}
\end{algorithm}

\subsection{Expressivity of Subgraph-based Aggregation}
In this subsection, we first define the subgraph-based neighborhood aggregation, and discuss the requirements of the subgraph feature map to achieve higher expressivity than 1-WL algorithm, then we show that KerGNN is one of the models that satisfy these requirements.

To leverage the structural information contained in the subgraph, we aggregate the subgraph information by finding a proper subgraph feature map $\Phi(G_v)$, and update the node representation of $v$ combining the subgraph feature map with $v$'s own feature map. Formally, we define this aggregation process as follows.

\textbf{Definition 4} (Subgraph-based aggregation). \textit{The graph neural network at layer $l$ deploying subgraph-based neighborhood aggregation updates feature mapping $\phi$ according to $\phi_l(v)=u\left(\phi_{l-1}(v),f\left(\Phi_{l-1}(G_v)\right)\right)$, where $u$ and $f$ are update and aggregation functions, respectively.}

GNNs distinguish different graphs by mapping them to different embeddings, which resembles the graph isomorphism test. \citet{xu2018powerful} characterize the representational capacity of MPNNs using the WL graph isomorphism test criterion, and show that MPNNs can be as powerful as 1-WL graph isomorphism test if the node update, aggregation, and graph-level readout function are injective. We follow the similar approach and show in the following that subgraph-based GNNs like KerGNNs can be at least as powerful as the 1-WL graph isomorphism test.

Because we are comparing the model's expressivity with the 1-WL algorithm which updates node labels based on the multiset of neighboring nodes, to achieve high expressivity, it is natural to think that $\Phi(G_v)$ should have a one-to-one relationship with respect to the multiset of nodes that subgraph $G_v$ contains. We show in Lemma \ref{lma:1} that the graph feature map induced by the random walk kernel satisfies this condition.

\begin{lemma}\label{lma:1}
if $\Phi(G)$ is the feature map of graph $G$ induced by the random walk graph kernel, then $\Phi(G)$ is injective with respect to the multiset of all its contained nodes $\{\!\!\{a(v): v\in\mathcal{V}(G)\}\!\!\}$, where $\{\!\!\{\cdot\}\!\!\}$ denotes the multiset and $a(v)$ is the label or attribute of node $v$.
\end{lemma}

The proof follows directly from the random walk kernel definition in \citet{gartner2003graph}, and we notice that the graph feature map induced by the WL graph kernel also satisfies this lemma. Based on this injective relationship between multiset and subgraph feature map, we can compare the expressivity of the subgraph-based GNN and 1-WL graph isomorphism test using the following theorem.

\begin{theorem}\label{thm:1}
Let $\mathcal{A}: \mathcal{G}\to\mathbb{R}^{d}$ be a GNN with a sufficient number of GNN layers, if the following conditions hold at layer $l$:

a) $\mathcal{A}$ aggregates and updates node features iteratively with
$\phi_l(v)=u\left(\phi_{l-1}(v),f\left(\Phi_{l-1}(G_v)\right)\right)$, where function $u$ and $f$ are injective, and $\Phi_{l-1}$ is the feature mapping induced by the random walk kernel;

b) $\mathcal{A}$'s graph-level readout, which operates on the multiset of node features $\left\{\!\!\left\{\phi_l(v)\right\}\!\!\right\}$, is injective;

\noindent then $\mathcal{A}$ maps any graphs $G$ and $H$ that 1-WL test decides as non-isomorphic to different embeddings, and there exist graph $G$ and $H$ that 1-WL test decides as isomorphic, but can be mapped to different embeddings by $\mathcal{A}$.
\end{theorem}

The proof is shown in Appendix. This theorem shows that subgraph-based GNNs can be more expressive than the 1-WL isomorphism test and thus MPNNs. In the KerGNN model, we do not explicitly calculate the subgraph feature map $\Phi(G_v)$ which lives in the high-dimensional space. Instead, we apply the kernel trick and use the subgraph feature map as $K(G_v,H) = \left< \Phi({G_v}) , \Phi({H}) \right>$. Then, the graph kernel function $K(\cdot,H)$ can be seen as a composition of functions $u$ and $f$. Therefore, according to Theorem \ref{thm:1}, to achieve high representational power, the graph kernel function needs to be injective with respect to the subgraph feature map $\Phi(G_v)$, and we introduce the following lemma to show that the KerGNN model satisfies this requirement. 

\textbf{Lemma 2} \textit{There exists a feature map $\Phi(H)$ so that $K(H,G_v) =\left<\Phi(H),\Phi(G_v)\right>$ is unique for different $\Phi(G_v)$. }

The proof is shown in Appendix. Besides, as shown in the definition of graph filters, in the KerGNN model we parameterize the node feature and adjacency matrix of graph filter $H$ instead of directly parameterizing $\Phi(H)$.

\subsection{Connections to CNNs}
\label{sec:inspiration}
Standard CNN models update the representation of each pixel by convolving filters with the patch centered at it, and in GNNs, a natural analog of the patch in the graph domain is the subgraph. While many MPNNs draw connections with CNNs by extending 2-D convolution to the graph convolution, we show in this subsection that both 2-D image convolution and KerGNN aggregation process can be viewed as applying kernel tricks to the input image or graph, and therefore, the KerGNN model naturally extends the CNN architecture into the graph domain, from a new kernel perspective. 

We first show in Appendix that under suitable assumptions, the 2-D image convolution can be viewed as applying kernel functions between input patches and filters. The basic idea is that we can rethink the 2-D convolution as projecting the input image patch into the kernel-induced Hilbert space. The projection is done by performing inner product between the patch and basis vectors, which can be calculated using the kernel trick, and the projected representation in the output space will be the output of the CNN layer. 

Then we can extend the same philosophy to the graph domain, by introducing subgraphs and topology-aware graph filters as the counterpart of patches and filters in CNNs, and KerGNN will adopt the kernel trick to project the input subgraph representation into the output space (detailed in Appendix). Based on these two observations, we can see that KerGNNs generalize CNNs into the graph domain by replacing the kernel function for vectors with the graph kernel function, which provides a new insight into designing GNN architecture, different from the spatial and spectral convolution perspectives.

\subsection{Connections to Existing GNNs}
As the subgraph of one node can be a more fruitful source of information than just the multiset of its neighbors, we show in this subsection that KerGNNs can generalize the standard MPNNs. From the point of view of KerGNNs, MPNNs deploy a simple graph filter with one node, and an appropriate kernel function can be chosen within KerGNN framework, such that KerGNNs iteratively update nodes' representations using neighborhood multiset aggregation like in MPNNs. For example, we show in Appendix that the node update rule of Graph Convolutional Network (GCN) \cite{kipf2016semi} can be treated as using one-node graph filters with properly-defined $\mathcal{R}$-convolution graph kernel. Our model generalizes most MPNN structures by deploying more complex graph filters with multiple nodes and learnable adjacency matrix, and using more expressive and efficient graph kernels.

\subsection{Time Complexity Analysis}
\label{sec:TimeAnalysis}
Most MPNNs incur a time complexity of $\mathcal{O}(n^2)$, or $\mathcal{O}(m)$ if the adjacency matrix is sparse containing $m$ non-zero entries, because updating the embedding of node $v$ involves $n_v$ neighbors, where $n_v$ is the degree of node $v$. In KerGNNs, we apply graph kernel with the subgraph $G_v$ instead of the whole graph, so the computational complexity would be related to the complexity of each subgraph. For the subgraph $G_v$ with $n_v+1$ nodes and adjacency matrix with $m_v$ non-zero entries, we update the representation of node $v$ by calculating the random walk kernel with Equation \ref{eq:kroneck2} in Appendix. This calculation takes a computation time of $\mathcal{O}(Pd(d'n_{GF}(n_{GF}+n_v+1)+m_v))$, where $P$ is the maximum length of the random walk, $d$ and $d'$ are the node dimensions of the current layer and next layer,  $n_{GF}$ is the number of nodes in each graph filter. In an undirected subgraph, $m_v$ represents the number of edges and will be greater than $n_v$ and smaller than $n_v(n_v-1)/2$. If we sum up the computation time for all the nodes in the entire graph, the time complexity of KerGNNs will range between $\mathcal{O}(n^2)$ and worst-case scenario (fully-connected graph) $\mathcal{O}(n^3)$. We experimentally compare the running time of the proposed model with several GNN benchmarks. As shown in Table \ref{table-time} in Appendix, KerGNNs achieve better or similar running time compared to the fastest benchmark method, and much less running time than higher-order GNNs.

\vspace{-0.1em}
\section{Experiments}
We evaluate the proposed model on graph classification task and node classification task (discussed in Appendix), and we also show the model interpretability by visualizing the graph filters in the trained models as well as the output graphs.

\begin{table*}[ht]
\caption{\textbf{Test set classification accuracies (\%)}. The mean accuracy and standard deviation are reported. Best performances are highlighted in bold. OOR means Out of Resources, either time or GPU memory.\vspace{-0.2in}}
\label{table}
\begin{center}
\begin{small}
\begin{sc}
\resizebox{\textwidth}{!}{
\begin{tabular}{lcccccccc}
\toprule
 & DD & NCI1 & PROTEINS & ENZYMES & IMDB-B & IMDB-M & REDDIT-B & COLLAB \\
\midrule
\# graphs        & 1178 &4110 &1113 &600 &1000 &1500 & 2000 &5000\\
\# classes       & 2 & 2 &2 &6 &2 &3 &2 &3 \\
avg. \# nodes  & 284 &30 &39 &33 &20 &13 &430 &74 \\
\midrule
SP   & 78.7$\pm$3.8 & 66.3$\pm$2.6 & 71.9$\pm$6.1 & 25.0$\pm$5.6 & 57.5$\pm$5.4 & 40.5$\pm$2.8 & 75.5$\pm$2.1 & 58.4$\pm$1.3 \\
PK   & 78.0$\pm$3.8 & 72.3$\pm$2.8 & 59.7$\pm$0.3 & 61.0$\pm$6.7 & 73.9$\pm$4.3 & 51.1$\pm$5.8 & 68.5$\pm$2.9 & 77.3$\pm$2.4 \\
WL-sub   & 77.5$\pm$3.5 & 79.5$\pm$3.3 & 74.8$\pm$3.2 & 51.2$\pm$5.3 & 72.5$\pm$4.6 & 51.5$\pm$5.8 & 67.2$\pm$4.2 & \textbf{77.5$\pm$2.4} \\
GNTK & OOR & \textbf{83.5$\pm$1.2} & 75.5$\pm$2.2 & 48.2$\pm$2.4 & \textbf{75.9$\pm$3.1} & \textbf{52.2$\pm$4.2} & OOR & OOR \\
\midrule
DGCNN      & 76.6$\pm$4.3 & 76.4$\pm$1.7 & 72.9$\pm$3.5 & 38.9$\pm$5.7 & 69.2$\pm$3.0 & 45.6$\pm$3.4 & 87.8$\pm$2.5 & 71.2$\pm$1.9 \\
DiffPool   & 75.0$\pm$3.5 & 76.9$\pm$1.9 & 73.7$\pm$3.5 & 59.5$\pm$5.6 & 68.4$\pm$3.3 & 45.6$\pm$3.4 & 89.1$\pm$1.6 & 68.9$\pm$2.0 \\
ECC        & 72.6$\pm$4.1 & 76.2$\pm$1.4 & 72.3$\pm$3.4 & 29.5$\pm$8.2 & 67.7$\pm$2.8 & 43.5$\pm$3.1 &  OOR & OOR \\
GIN        & 75.3$\pm$2.9 & 80.0$\pm$1.4 & 73.3$\pm$4.0 & 59.6$\pm$4.5 & 71.2$\pm$3.9 & 48.5$\pm$3.3 & 89.9$\pm$1.9 & 75.6$\pm$2.3 \\
GraphSAGE  & 72.9$\pm$2.0 & 76.0$\pm$1.8 & 73.0$\pm$4.5 & 58.2$\pm$6.0 & 68.8$\pm$4.5 & 47.6$\pm$3.5 & 84.3$\pm$1.9 & 73.9$\pm$1.7 \\
\midrule
RWGNN  & 77.6$\pm$4.7 & 73.9$\pm$1.3 & 74.7$\pm$3.3 & 57.6$\pm$6.3 & 70.8$\pm$4.8 & 48.8$\pm$2.9 & 90.4$\pm$1.9 & 71.9$\pm$2.5 \\
GCKN  & 77.3$\pm$4.0 & 79.2$\pm$1.2 & 76.1$\pm$2.8 & 59.3$\pm$5.6 & 74.5$\pm$1.2 & 51.0$\pm$3.9 & OOR& 74.3$\pm$2.8 \\
1-2-3 GNN  & OOR & 72.7$\pm$2.9 & 74.5$\pm$5.6 & OOR & 70.7$\pm$3.4 & 50.2$\pm$2.2 & \textbf{91.1$\pm$2.1} & OOR \\
Powerful GNN & OOR & 83.4$\pm$1.8 & 75.9$\pm$3.3 & 54.8$\pm$5.5 & 73.0$\pm$4.9 & 50.5$\pm$3.2 & OOR & 75.4$\pm$1.4 \\
\midrule
KerGNN-1  &  77.6$\pm$3.7 &  74.3$\pm$2.2 &  75.8$\pm$3.5 &  \textbf{62.1$\pm$5.5} &  74.4$\pm$4.3 &  51.6$\pm$3.1 &  81.5$\pm$1.9 &  70.5$\pm$1.6\\
KerGNN-2 &  \textbf{78.9$\pm$3.5}  &  76.3$\pm$2.6 &  75.5$\pm$4.6 &  55.0$\pm$5.0  &  73.7$\pm$4.0 &  50.9$\pm$5.1 &  82.0$\pm$2.5 &  72.7$\pm$2.1\\
KerGNN-3 &  75.5$\pm$3.1 &  80.5$\pm$1.9 &  \textbf{76.5$\pm$3.9} &  54.1$\pm$4.3 & 72.1$\pm$4.6 &  50.1$\pm$4.5 &  82.0$\pm$1.9&  71.1$\pm$2.0\\
KerGNN-2-DRW  & 77.0$\pm$4.4 & 82.8$\pm$1.8 &  76.1$\pm$4.1 &  59.5$\pm$4.5 & 71.1$\pm$4.1 & 50.5$\pm$3.1 & 89.5$\pm$1.6 & 75.1$\pm$2.3\\

\bottomrule
\end{tabular}}
\end{sc}
\end{small}
\end{center}
\vspace{-20pt}
\end{table*}
\subsubsection{Experiment Settings}
\subsubsection{Datasets.}
We evaluate our proposed KerGNN model on 8 publicly available graph classification datasets. Specifically, we use DD \cite{dobson2003distinguishing}, PROTEINS \cite{borgwardt2005protein}, NCI1 \cite{schomburg2004brenda}, ENZYMES \cite{schomburg2004brenda} for binary and multi-class classification of biological and chemical compounds, and we also use the social datasets IMDB-BINARY, IMDB-MULTI, REDDIT-BINARY, and COLLAB \cite{yanardag2015deep}.

\subsubsection{Setup.}
To make a fair comparison with state-of-the-art GNNs, we follow the cross-validation procedure described in \citet{errica2019fair}. We use a 10-fold cross-validation for model assessment and an inner holdout technique with a 90\%/10\% training/validation split for model selection, following the same dataset index splits as \citet{errica2019fair}. 
Besides, we use Adam optimizer with an initial learning rate of 0.01 and decay the learning rate by half in every 50 epochs. For the four social datasets, we use node degrees as the input attributes for each node, and for the four bio/chemical datasets, we use node labels or attributes as the input feature for each node.
\vspace{-0.2em}
\subsubsection{Hyper-parameters.}
The hyper-parameters that we tune for each dataset include the learning rate, the dropout rate, the number of layers of KerGNNs and MLP, the number of graph filters at each layer, the number of nodes in each graph filter, the number of nodes for each subgraph, and the hidden dimension of each KerGNN layer. For the random walk kernel, we also tune the length of random walks.
\vspace{-0.2em}
\subsubsection{Baseline Models.}
We consider the KerGNN model with single and multiple KerGNN layers, namely KerGNN-$L$, corresponding to KerGNN model with $L$ layers, and KerGNN-$L$-DRW representing the model deploying the deep random walk kernel.
We also compare our models with widely-used GNNs: DGCNN \cite{zhang2018end}, DiffPool \cite{ying2018hierarchical}, ECC \cite{simonovsky2017dynamic}, GIN \cite{xu2018powerful}, GraphSAGE \cite{hamilton2017inductive}, RWGNN \cite{rwgnn}, GCKN \cite{chen2020convolutional}, and two high-order GNNs: 1-2-3 GNN \cite{morris2019weisfeiler} and Powerful GNN \cite{maron2019provably}. Part of the results for these baseline GNNs are taken from \citet{errica2019fair}, and we run GCKN, 1-2-3 GNN and Powerful GNN using the official implementations. In addition, we also compare the proposed KerGNN model with three popular GNN-unrelated graph kernels: shortest path (SP) kernel \cite{borgwardt2005shortest}, propagation (PK) kernel \cite{neumann2016propagation}, the Weisfeiler-Lehman subtree (WL-sub) kernel \cite{shervashidze2011weisfeiler} and GNN-related GNTK \cite{du2019graph}. We use the GraKeL library \cite{JMLR:v21:18-370} to implement these graph kernels and run GNTK using the official implementation.

\vspace{-0.3em}
\subsection{Results}
\vspace{-0.1em}

The graph classification results are shown in Table \ref{table}, with the best results highlighted in bold. We can see that the proposed models achieve superior performance than conventional GNNs with 1-WL limits, and achieve similar performance compared with high-order GNNs, with less running time. The single-layer KerGNN model performs well on small graphs like IMDB social datasets. For larger graphs, deeper models with more layers or with deep random walk kernel perform better. We show more experimental results, model parameter studies, and node classification results in Appendix. The optimal parameters of the graph filter are different for different datasets, depending on the local structures of different types of graphs, e.g., the star patterns in graphs of REDDIT-B and the ring and chain patterns in graphs of NCI1.

\subsection{Model Interpretability}

\begin{figure}[tb]
  \centering \includegraphics[width=1\columnwidth]{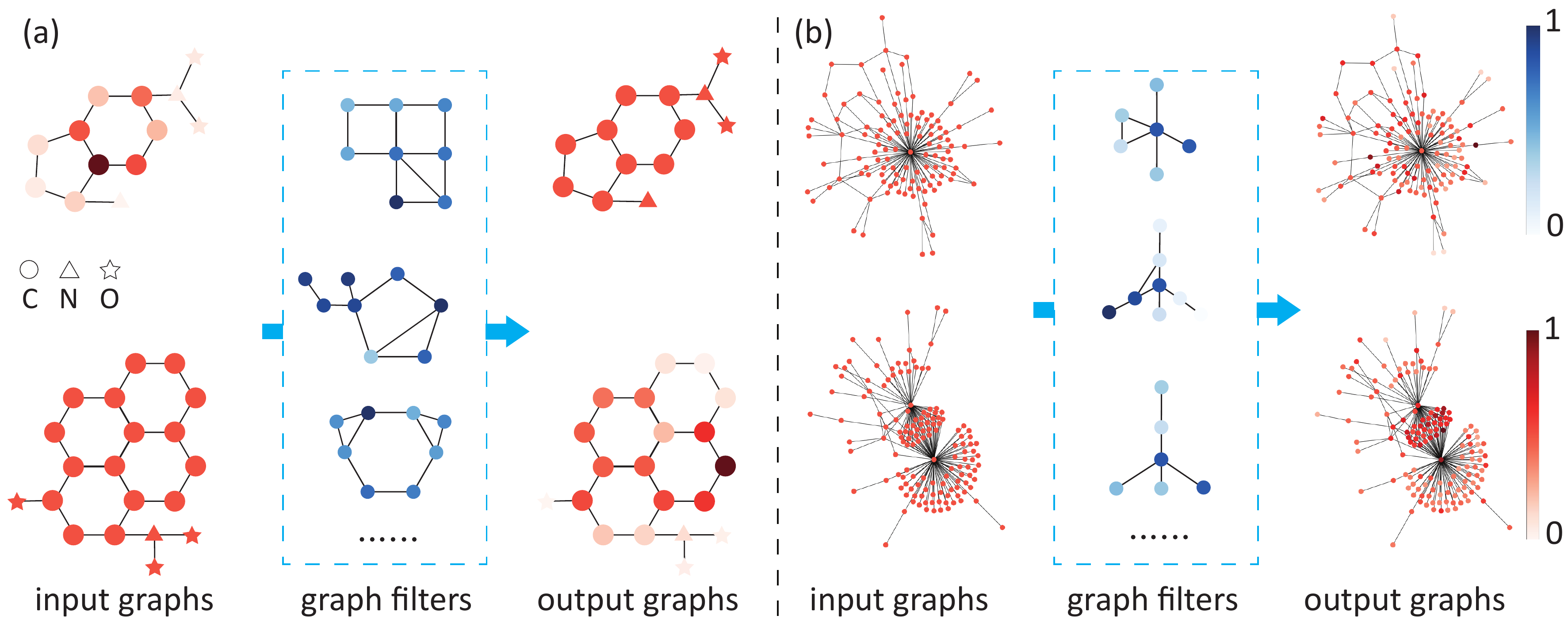}
  \caption{\textbf{Model visualization}. Input graphs are drawn from (a) MUTAG and (b) REDDIT-B datasets with different node shapes corresponding to different atom types. In both graph filters and output graphs, node color represents relative attribute value.}
  \label{fig:visual}
  \vspace{-0.2in}
\end{figure}

Visualizing the filters in CNNs gives insights into what features CNNs focus on. Following the same idea, we can also visualize the trained graph filters, which indicate some key structures of the input dataset. We visualize the graph filters trained with MUTAG \footnote{We use the MUTAG dataset for visualization due to its easily interpretable structure. However, we do not use this dataset in cross-validation because its number of graphs is too small.} \cite{KKMMN2016} and REDDIT-B dataset in Figure \ref{fig:visual}. The MUTAG dataset consists of 188 chemical compounds divided into two classes according to their mutagenic effect on a bacterium. As shown in the input graphs in Figure \ref{fig:visual}(a), most of the MUTAG graphs in the dataset consist of ring structures with 6 carbon atoms.

KerGNNs and other standard GNNs can generate output graphs with updated node attributes, and we can extract important nodes for the classification tasks using relative attribute values, which is shown
in output graphs in Figure \ref{fig:visual}. We can make several observations from the output MUTAG chemical structures: 1) The carbon atoms at the connection points of rings are more important than those connected with atom groups, which are more important than those at the remaining positions. 2) The atoms in the atom group are always less important than those carbon atoms in the carbon ring.

Compared to standard GNN variants, KerGNNs have graph filters as extra information to help explain the predictions of the model. To visualize the graph filters, we extract the adjacency matrix and the attribute matrix for each graph filter from the trained KerGNN layer. We then adopt the \textit{ReLU} functions to prune the unimportant edges. 
In Figure \ref{fig:visual}, we use different sizes of nodes to denote the relative importance of nodes. For the MUTAG dataset, we can see most of the graph filters have ring structures, similar to the carbon rings at the input graphs, and some graph filters have small connected rings, similar to the concatenated carbon rings. It should be noted that the number of nodes in the rings of graph filters may not be equal to 6 because we limit the total number of nodes to be 8. KerGNN layers utilize these rings in the graph filter to match against the local structural patterns (e.g., carbon rings) in the input graphs, using the graph kernels. This indicates the importance of carbon rings in the mutagenic effect, which also corresponds to our observations in the output graphs.

\section{Conclusion}

In this paper, we have proposed Kernel Graph Neural Networks (KerGNNs), a new graph neural network framework that is not restricted to the theoretical limits of the message passing aggregation. KerGNNs are inspired by several characteristics of CNNs and can be seen as a natural extension of CNNs in the graph domain, from the viewpoint of the kernel methods. KerGNNs achieve competitive performance on a variety of datasets compared with several GNNs and graph kernels, and can provide improved explainability and transparency by visualizing the graph filters and output graphs.

\section{Acknowledgment}

This work was partially supported by the U.S. Office of Naval Research under
Grant N00173-21-1-G006, the U.S. National Science Foundation AI Institute Athena under Grant CNS-2112562, and the U.S. Army Research Laboratory and the U.K. Ministry of Defence under Agreement Number W911NF-16-3-0001. The views and conclusions contained in this document are those of the authors and should not be interpreted as representing the official policies, either expressed or implied, of U.S. Office of Naval Research, the U.S. National Science Foundation, the U.S. Army Research Laboratory, the U.S. Government, the U.K. Ministry of Defence or the U.K. Government. The U.S. and U.K. Governments are authorized to reproduce and distribute reprints for Government purposes notwithstanding any copyright notation hereon. 


\bibliography{kergnn}

\begin{thebibliography}{50}
\providecommand{\natexlab}[1]{#1}

\bibitem[{Abu-El-Haija et~al.(2019)Abu-El-Haija, Perozzi, Kapoor, Alipourfard,
  Lerman, Harutyunyan, Ver~Steeg, and Galstyan}]{abu2019mixhop}
Abu-El-Haija, S.; Perozzi, B.; Kapoor, A.; Alipourfard, N.; Lerman, K.;
  Harutyunyan, H.; Ver~Steeg, G.; and Galstyan, A. 2019.
\newblock Mixhop: Higher-order graph convolutional architectures via sparsified
  neighborhood mixing.
\newblock In \emph{international conference on machine learning}, 21--29. PMLR.

\bibitem[{Baldassarre and Azizpour(2019)}]{baldassarre2019explainability}
Baldassarre, F.; and Azizpour, H. 2019.
\newblock Explainability techniques for graph convolutional networks.
\newblock \emph{arXiv preprint arXiv:1905.13686}.

\bibitem[{Borgwardt and Kriegel(2005)}]{borgwardt2005shortest}
Borgwardt, K.~M.; and Kriegel, H.-P. 2005.
\newblock Shortest-path kernels on graphs.
\newblock In \emph{Fifth IEEE international conference on data mining
  (ICDM'05)}, 8--pp. IEEE.

\bibitem[{Borgwardt et~al.(2005)Borgwardt, Ong, Sch{\"o}nauer, Vishwanathan,
  Smola, and Kriegel}]{borgwardt2005protein}
Borgwardt, K.~M.; Ong, C.~S.; Sch{\"o}nauer, S.; Vishwanathan, S.; Smola,
  A.~J.; and Kriegel, H.-P. 2005.
\newblock Protein function prediction via graph kernels.
\newblock \emph{Bioinformatics}, 21(suppl\_1): i47--i56.

\bibitem[{Chen, Jacob, and Mairal(2020)}]{chen2020convolutional}
Chen, D.; Jacob, L.; and Mairal, J. 2020.
\newblock Convolutional kernel networks for graph-structured data.
\newblock In \emph{International Conference on Machine Learning}, 1576--1586.
  PMLR.

\bibitem[{Chen et~al.(2020)Chen, Wei, Huang, Ding, and Li}]{chen2020simple}
Chen, M.; Wei, Z.; Huang, Z.; Ding, B.; and Li, Y. 2020.
\newblock Simple and deep graph convolutional networks.
\newblock In \emph{International Conference on Machine Learning}, 1725--1735.
  PMLR.

\bibitem[{Chen, Li, and Bruna(2017)}]{chen2017supervised}
Chen, Z.; Li, X.; and Bruna, J. 2017.
\newblock Supervised community detection with line graph neural networks.
\newblock \emph{arXiv preprint arXiv:1705.08415}.

\bibitem[{Dobson and Doig(2003)}]{dobson2003distinguishing}
Dobson, P.~D.; and Doig, A.~J. 2003.
\newblock Distinguishing enzyme structures from non-enzymes without alignments.
\newblock \emph{Journal of molecular biology}, 330(4): 771--783.

\bibitem[{Du et~al.(2019)Du, Hou, P{\'o}czos, Salakhutdinov, Wang, and
  Xu}]{du2019graph}
Du, S.~S.; Hou, K.; P{\'o}czos, B.; Salakhutdinov, R.; Wang, R.; and Xu, K.
  2019.
\newblock Graph neural tangent kernel: Fusing graph neural networks with graph
  kernels.
\newblock \emph{arXiv preprint arXiv:1905.13192}.

\bibitem[{Errica et~al.(2019)Errica, Podda, Bacciu, and
  Micheli}]{errica2019fair}
Errica, F.; Podda, M.; Bacciu, D.; and Micheli, A. 2019.
\newblock A fair comparison of graph neural networks for graph classification.
\newblock \emph{arXiv preprint arXiv:1912.09893}.

\bibitem[{G{\"a}rtner, Flach, and Wrobel(2003)}]{gartner2003graph}
G{\"a}rtner, T.; Flach, P.; and Wrobel, S. 2003.
\newblock On graph kernels: Hardness results and efficient alternatives.
\newblock In \emph{Learning theory and kernel machines}, 129--143. Springer.

\bibitem[{Gilmer et~al.(2017)Gilmer, Schoenholz, Riley, Vinyals, and
  Dahl}]{gilmer2017neural}
Gilmer, J.; Schoenholz, S.~S.; Riley, P.~F.; Vinyals, O.; and Dahl, G.~E. 2017.
\newblock Neural message passing for quantum chemistry.
\newblock In \emph{International Conference on Machine Learning}, 1263--1272.
  PMLR.

\bibitem[{Hamilton, Ying, and Leskovec(2017)}]{hamilton2017inductive}
Hamilton, W.~L.; Ying, R.; and Leskovec, J. 2017.
\newblock Inductive representation learning on large graphs.
\newblock \emph{arXiv preprint arXiv:1706.02216}.

\bibitem[{Haussler(1999)}]{haussler1999convolution}
Haussler, D. 1999.
\newblock Convolution kernels on discrete structures.
\newblock Technical report, Technical report, Department of Computer Science,
  University of California~….

\bibitem[{Kashima, Tsuda, and Inokuchi(2003)}]{kashima2003marginalized}
Kashima, H.; Tsuda, K.; and Inokuchi, A. 2003.
\newblock Marginalized kernels between labeled graphs.
\newblock In \emph{Proceedings of the 20th international conference on machine
  learning (ICML-03)}, 321--328.

\bibitem[{Kersting et~al.(2016)Kersting, Kriege, Morris, Mutzel, and
  Neumann}]{KKMMN2016}
Kersting, K.; Kriege, N.~M.; Morris, C.; Mutzel, P.; and Neumann, M. 2016.
\newblock Benchmark Data Sets for Graph Kernels.

\bibitem[{Kipf and Welling(2016)}]{kipf2016semi}
Kipf, T.~N.; and Welling, M. 2016.
\newblock Semi-supervised classification with graph convolutional networks.
\newblock \emph{arXiv preprint arXiv:1609.02907}.

\bibitem[{Klicpera, Bojchevski, and G{\"u}nnemann(2018)}]{klicpera2018predict}
Klicpera, J.; Bojchevski, A.; and G{\"u}nnemann, S. 2018.
\newblock Predict then propagate: Graph neural networks meet personalized
  pagerank.
\newblock \emph{arXiv preprint arXiv:1810.05997}.

\bibitem[{Klicpera, Gro{\ss}, and
  G{\"u}nnemann(2020)}]{klicpera2020directional}
Klicpera, J.; Gro{\ss}, J.; and G{\"u}nnemann, S. 2020.
\newblock Directional message passing for molecular graphs.
\newblock \emph{arXiv preprint arXiv:2003.03123}.

\bibitem[{Kriege, Johansson, and Morris(2020)}]{kriege2020survey}
Kriege, N.~M.; Johansson, F.~D.; and Morris, C. 2020.
\newblock A survey on graph kernels.
\newblock \emph{Applied Network Science}, 5(1): 1--42.

\bibitem[{Lei et~al.(2017)Lei, Jin, Barzilay, and Jaakkola}]{lei2017deriving}
Lei, T.; Jin, W.; Barzilay, R.; and Jaakkola, T. 2017.
\newblock Deriving neural architectures from sequence and graph kernels.
\newblock In \emph{International Conference on Machine Learning}, 2024--2033.
  PMLR.

\bibitem[{Leman and Weisfeiler(1968)}]{leman1968reduction}
Leman, A.; and Weisfeiler, B. 1968.
\newblock A reduction of a graph to a canonical form and an algebra arising
  during this reduction.
\newblock \emph{Nauchno-Technicheskaya Informatsiya}, 2(9): 12--16.

\bibitem[{Mairal(2016)}]{mairal2016end}
Mairal, J. 2016.
\newblock End-to-end kernel learning with supervised convolutional kernel
  networks.
\newblock \emph{arXiv preprint arXiv:1605.06265}.

\bibitem[{Mairal et~al.(2014)Mairal, Koniusz, Harchaoui, and
  Schmid}]{mairal2014convolutional}
Mairal, J.; Koniusz, P.; Harchaoui, Z.; and Schmid, C. 2014.
\newblock Convolutional kernel networks.
\newblock \emph{arXiv preprint arXiv:1406.3332}.

\bibitem[{Maron et~al.(2019)Maron, Ben-Hamu, Serviansky, and
  Lipman}]{maron2019provably}
Maron, H.; Ben-Hamu, H.; Serviansky, H.; and Lipman, Y. 2019.
\newblock Provably powerful graph networks.
\newblock \emph{arXiv preprint arXiv:1905.11136}.

\bibitem[{Morris et~al.(2019)Morris, Ritzert, Fey, Hamilton, Lenssen, Rattan,
  and Grohe}]{morris2019weisfeiler}
Morris, C.; Ritzert, M.; Fey, M.; Hamilton, W.~L.; Lenssen, J.~E.; Rattan, G.;
  and Grohe, M. 2019.
\newblock Weisfeiler and leman go neural: Higher-order graph neural networks.
\newblock In \emph{Proceedings of the AAAI Conference on Artificial
  Intelligence}, volume~33, 4602--4609.

\bibitem[{Navarin, Tran, and Sperduti(2018)}]{navarin2018pre}
Navarin, N.; Tran, D.~V.; and Sperduti, A. 2018.
\newblock Pre-training graph neural networks with kernels.
\newblock \emph{arXiv preprint arXiv:1811.06930}.

\bibitem[{Neumann et~al.(2016)Neumann, Garnett, Bauckhage, and
  Kersting}]{neumann2016propagation}
Neumann, M.; Garnett, R.; Bauckhage, C.; and Kersting, K. 2016.
\newblock Propagation kernels: efficient graph kernels from propagated
  information.
\newblock \emph{Machine Learning}, 102(2): 209--245.

\bibitem[{Nguyen and Maehara(2020)}]{nguyen2020graph}
Nguyen, H.; and Maehara, T. 2020.
\newblock Graph Homomorphism Convolution.
\newblock In \emph{International Conference on Machine Learning}, 7306--7316.
  PMLR.

\bibitem[{Nikolentzos et~al.(2018)Nikolentzos, Meladianos, Tixier, Skianis, and
  Vazirgiannis}]{nikolentzos2018kernel}
Nikolentzos, G.; Meladianos, P.; Tixier, A. J.-P.; Skianis, K.; and
  Vazirgiannis, M. 2018.
\newblock Kernel graph convolutional neural networks.
\newblock In \emph{International Conference on Artificial Neural Networks},
  22--32. Springer.

\bibitem[{Nikolentzos and Vazirgiannis(2020)}]{rwgnn}
Nikolentzos, G.; and Vazirgiannis, M. 2020.
\newblock Random Walk Graph Neural Networks.
\newblock In \emph{Conference on Neural Information Processing System}. PMLR.

\bibitem[{Pei et~al.(2020)Pei, Wei, Chang, Lei, and Yang}]{pei2020geom}
Pei, H.; Wei, B.; Chang, K. C.-C.; Lei, Y.; and Yang, B. 2020.
\newblock Geom-gcn: Geometric graph convolutional networks.
\newblock \emph{arXiv preprint arXiv:2002.05287}.

\bibitem[{Pope et~al.(2019)Pope, Kolouri, Rostami, Martin, and
  Hoffmann}]{pope2019explainability}
Pope, P.~E.; Kolouri, S.; Rostami, M.; Martin, C.~E.; and Hoffmann, H. 2019.
\newblock Explainability methods for graph convolutional neural networks.
\newblock In \emph{Proceedings of the IEEE/CVF Conference on Computer Vision
  and Pattern Recognition}, 10772--10781.

\bibitem[{Rozemberczki, Allen, and Sarkar(2021)}]{rozemberczki2021multi}
Rozemberczki, B.; Allen, C.; and Sarkar, R. 2021.
\newblock Multi-scale attributed node embedding.
\newblock \emph{Journal of Complex Networks}, 9(2): cnab014.

\bibitem[{Sato, Yamada, and Kashima(2019)}]{sato2019approximation}
Sato, R.; Yamada, M.; and Kashima, H. 2019.
\newblock Approximation ratios of graph neural networks for combinatorial
  problems.
\newblock \emph{arXiv preprint arXiv:1905.10261}.

\bibitem[{Schomburg et~al.(2004)Schomburg, Chang, Ebeling, Gremse, Heldt, Huhn,
  and Schomburg}]{schomburg2004brenda}
Schomburg, I.; Chang, A.; Ebeling, C.; Gremse, M.; Heldt, C.; Huhn, G.; and
  Schomburg, D. 2004.
\newblock BRENDA, the enzyme database: updates and major new developments.
\newblock \emph{Nucleic acids research}, 32(suppl\_1): D431--D433.

\bibitem[{Sen et~al.(2008)Sen, Namata, Bilgic, Getoor, Galligher, and
  Eliassi-Rad}]{sen2008collective}
Sen, P.; Namata, G.; Bilgic, M.; Getoor, L.; Galligher, B.; and Eliassi-Rad, T.
  2008.
\newblock Collective classification in network data.
\newblock \emph{AI magazine}, 29(3): 93--93.

\bibitem[{Shervashidze et~al.(2011)Shervashidze, Schweitzer, Van~Leeuwen,
  Mehlhorn, and Borgwardt}]{shervashidze2011weisfeiler}
Shervashidze, N.; Schweitzer, P.; Van~Leeuwen, E.~J.; Mehlhorn, K.; and
  Borgwardt, K.~M. 2011.
\newblock Weisfeiler-lehman graph kernels.
\newblock \emph{Journal of Machine Learning Research}, 12(9).

\bibitem[{Shervashidze et~al.(2009)Shervashidze, Vishwanathan, Petri, Mehlhorn,
  and Borgwardt}]{shervashidze2009efficient}
Shervashidze, N.; Vishwanathan, S.; Petri, T.; Mehlhorn, K.; and Borgwardt, K.
  2009.
\newblock Efficient graphlet kernels for large graph comparison.
\newblock In \emph{Artificial intelligence and statistics}, 488--495. PMLR.

\bibitem[{Siglidis et~al.(2020)Siglidis, Nikolentzos, Limnios, Giatsidis,
  Skianis, and Vazirgiannis}]{JMLR:v21:18-370}
Siglidis, G.; Nikolentzos, G.; Limnios, S.; Giatsidis, C.; Skianis, K.; and
  Vazirgiannis, M. 2020.
\newblock GraKeL: A Graph Kernel Library in Python.
\newblock \emph{Journal of Machine Learning Research}, 21(54): 1--5.

\bibitem[{Simonovsky and Komodakis(2017)}]{simonovsky2017dynamic}
Simonovsky, M.; and Komodakis, N. 2017.
\newblock Dynamic edge-conditioned filters in convolutional neural networks on
  graphs.
\newblock In \emph{Proceedings of the IEEE conference on computer vision and
  pattern recognition}, 3693--3702.

\bibitem[{Veli{\v{c}}kovi{\'c} et~al.(2017)Veli{\v{c}}kovi{\'c}, Cucurull,
  Casanova, Romero, Lio, and Bengio}]{velivckovic2017graph}
Veli{\v{c}}kovi{\'c}, P.; Cucurull, G.; Casanova, A.; Romero, A.; Lio, P.; and
  Bengio, Y. 2017.
\newblock Graph attention networks.
\newblock \emph{arXiv preprint arXiv:1710.10903}.

\bibitem[{Vishwanathan et~al.(2006)Vishwanathan, Borgwardt, Schraudolph
  et~al.}]{vishwanathan2006fast}
Vishwanathan, S.; Borgwardt, K.~M.; Schraudolph, N.~N.; et~al. 2006.
\newblock Fast computation of graph kernels.
\newblock In \emph{NIPS}, volume~19, 131--138. Citeseer.

\bibitem[{Xu et~al.(2018{\natexlab{a}})Xu, Hu, Leskovec, and
  Jegelka}]{xu2018powerful}
Xu, K.; Hu, W.; Leskovec, J.; and Jegelka, S. 2018{\natexlab{a}}.
\newblock How powerful are graph neural networks?
\newblock \emph{arXiv preprint arXiv:1810.00826}.

\bibitem[{Xu et~al.(2018{\natexlab{b}})Xu, Li, Tian, Sonobe, Kawarabayashi, and
  Jegelka}]{xu2018representation}
Xu, K.; Li, C.; Tian, Y.; Sonobe, T.; Kawarabayashi, K.-i.; and Jegelka, S.
  2018{\natexlab{b}}.
\newblock Representation learning on graphs with jumping knowledge networks.
\newblock In \emph{International Conference on Machine Learning}, 5453--5462.
  PMLR.

\bibitem[{Yanardag and Vishwanathan(2015)}]{yanardag2015deep}
Yanardag, P.; and Vishwanathan, S. 2015.
\newblock Deep graph kernels.
\newblock In \emph{Proceedings of the 21th ACM SIGKDD international conference
  on knowledge discovery and data mining}, 1365--1374.

\bibitem[{Ying et~al.(2019)Ying, Bourgeois, You, Zitnik, and
  Leskovec}]{ying2019gnnexplainer}
Ying, R.; Bourgeois, D.; You, J.; Zitnik, M.; and Leskovec, J. 2019.
\newblock Gnnexplainer: Generating explanations for graph neural networks.
\newblock \emph{Advances in neural information processing systems}, 32: 9240.

\bibitem[{Ying et~al.(2018)Ying, You, Morris, Ren, Hamilton, and
  Leskovec}]{ying2018hierarchical}
Ying, R.; You, J.; Morris, C.; Ren, X.; Hamilton, W.~L.; and Leskovec, J. 2018.
\newblock Hierarchical graph representation learning with differentiable
  pooling.
\newblock \emph{arXiv preprint arXiv:1806.08804}.

\bibitem[{Zhang and Chen(2018)}]{zhang2018link}
Zhang, M.; and Chen, Y. 2018.
\newblock Link prediction based on graph neural networks.
\newblock \emph{arXiv preprint arXiv:1802.09691}.

\bibitem[{Zhang et~al.(2018)Zhang, Cui, Neumann, and Chen}]{zhang2018end}
Zhang, M.; Cui, Z.; Neumann, M.; and Chen, Y. 2018.
\newblock An end-to-end deep learning architecture for graph classification.
\newblock In \emph{Proceedings of the AAAI Conference on Artificial
  Intelligence}, volume~32.

\end{thebibliography}

\appendix
\onecolumn
\onecolumn

\section{Calculation of Random Walk Kernel}

To calculate the random walk kernel between two graphs, we have to calculate Equation \ref{eq:ssa} in the main paper:
\begin{equation}
\label{eq:kroneck1}
\begin{split}
   K_p(G_1,G_2) = \mathbf{s}^T A_{\times}^{p} \mathbf{s},
\end{split}
\end{equation}
where $A_{\times}=A_2 \otimes A_1$, $\otimes$ denotes the Kronecker product between two matrices, $s=\text{vec}(\mathbf{S})$, and $\text{vec}(\cdot)$ denotes flattening a matrix into a vector by stacking all the columns. Besides $\mathbf{S}=\mathbf{X}_1 \mathbf{X}_2^T\in \mathbb{R}^{n_1\times n_2}$. In the following, we assume $G_1$ and $G_2$ are undirected graphs, which is the case for all the datasets we use in the experiments. According to the properties of Kronecker product 
\begin{equation}
\begin{split}
   A_{\times}^p=A_2^p \otimes A_1^p ,
\end{split}
\end{equation}
and 
\begin{equation}
\begin{split}
   \text{vec}(\mathbf{AXB}) = (\mathbf{B}^T\otimes \mathbf{A})\text{vec}(\mathbf{X}) ,
\end{split}
\end{equation}
we can calculate Equation \ref{eq:kroneck1} as 
\begin{equation}
\label{eq:kroneck2}
\begin{split}
   K_p(G_1,G_2) & = \mathbf{s}^T A_{\times}^{p} \mathbf{s}\\
   & = \mathbf{s}^T ({A_2^p \otimes A_1^p}) \mathbf{s}\\
   & = \text{vec}\left( \mathbf{X}_1\mathbf{X}_2^T\right)^T ({A_2^p \otimes A_1^p}){\text{vec}\left(\mathbf{X}_1\mathbf{X}_2^T\right)}\\
   & = \text{vec}\left( \mathbf{X}_1\mathbf{X}_2^T\right)^T \text{vec} \left((A_1^p)^T \mathbf{X}_1 \mathbf{X}_2^T A_2^p \right)\\
   & = \text{vec}\left( \mathbf{X}_1\mathbf{X}_2^T\right)^T \text{vec} \left(A_1^p \mathbf{X}_1 (A_2^p\mathbf{X}_2)^T \right)\\
   & =\sum_{i=1}^{n_1}\sum_{j=1}^{n_2} \left[\left( \mathbf{X}_1\mathbf{X}_2^T\right) \odot \left(A_1^p \mathbf{X}_1 (A_2^p\mathbf{X}_2)^T \right)\right]_{ij},\\
\end{split}
\end{equation}
where $\odot$ means Hadamard (element-wise) product,  In the KerGNN model, $G_1$ and $G_2$ represent the graph filter and the input graph, respectively, and we can use Equation \ref{eq:kroneck2} to avoid calculating the direct product graph.

\section{Proofs}
\subsection{Proof of Theorem 1}
Let $\mathcal{A}$ be a graph neural network which satisfies condition a) and b). We first prove the first part of conclusions that $\mathcal{A}$ can map any graphs $G$ and $H$ that 1-WL test decides as non-isomorphic to different embeddings. 
Suppose starting from iteration $l$, the 1-WL test decides $G$ and $H$ are non-isomorphic (before that, 1-WL test cannot distinguish two graphs), but graph neural network maps them to the same embeddings $\mathcal{A}(G)=\mathcal{A}(H)$. This indicates that $G$ and $H$ always have the same labels for iteration $i-1$ and $i$ for any $i=1,...,l-1$ in the 1-WL test. Next we hope to reach a contradiction to this statement. To find this contradiction, we first show on graph $G$ or $H$, if node features in the graph neural network $\phi_i(v_1)=\phi_i(v_2)$, we always have 1-WL node labels $a_i(v_1)=a_i(v_2)$ for any iteration $i$. This apparently holds for $i=0$ because 1-WL and graph neural network start with the same node features. Suppose this holds for iteration $j$, if for any $v_1$ and $v_2$, $\phi_{j+1}(v_1)=\phi_{j+1}(v_2)$, then according to the node update rule in Theorem \ref{thm:1}, we can get
\begin{equation}
u\left(\phi_{j}(v_1),f\left(\Phi_{j}(G_{v_1})\right)\right)=u\left(\phi_{j}(v_2),f\left(\Phi_{j}(G_{v_2})\right)\right).
\end{equation}
Because $u$ and $f$ are both injective, we then obtain
\begin{equation}
\left(\phi_{j}(v_1),\Phi_{j}(G_{v_1})\right)=\left(\phi_{j}(v_2),\Phi_{j}(G_{v_2})\right).
\end{equation}
According to Lemma 1, if $\Phi_{j}(G_{v_1})=\Phi_{j}(G_{v_2})$, then $\{\!\!\{ \phi_{j}(w),w\in \mathcal{V}(G_{v_1})\}\!\!\}=\{\!\!\{ \phi_{j}(w),w\in \mathcal{V}(G_{v_2})\}\!\!\}$, and because $\phi_{j}(v_1)=\phi_{j}(v_2)$, we can get
\begin{equation}
\left(\phi_{j}(v_1),\{\!\!\{ \phi_{j}(w),w\in \mathcal{N}({v_1})\}\!\!\}\right)=\left(\phi_{j}(v_2),\{\!\!\{ \phi_{j}(w),w\in \mathcal{N}({v_2})\}\!\!\}\right).
\end{equation}
By our assumption at iteration $j$, we must have
\begin{equation}
\left(a_{j}(v_1),\{\!\!\{a_{j}(w),w\in \mathcal{N}({v_1})\}\!\!\}\right)=\left(a_{j}(v_2),\{\!\!\{ a_{j}(w),w\in \mathcal{N}({v_2})\}\!\!\}\right).
\end{equation}
Because the mapping in 1-WL test is injective with respect to the node label and the multiset of neighborhood labels, we get $a_{j+1}(v_1)=a_{j+1}(v_2)$. By induction, if node features in the graph neural network $\phi_i(v_1)=\phi_i(v_2)$, we always have 1-WL node labels $a_i(v_1)=a_i(v_2)$ for any iteration $i$. This creates a valid mapping $q$ such that $a_i(v)=q(\phi_i(v))$ for any node $v$ in the graph.

Because 1-WL decides graphs $G$ and $H$ as non-isomorphic, which means $\{\!\!\{a_l(v),v\in\mathcal{V}(G)\}\!\!\}\neq \{\!\!\{a_l(v),v\in\mathcal{V}(H)\}\!\!\}$, at layer $l$. With the mapping between $a_l(v)$ and $\phi_l(v)$, we can get $\{\!\!\{\phi_l(v),v\in\mathcal{V}(G)\}\!\!\}\neq \{\!\!\{\phi_l(v),v\in\mathcal{V}(H)\}\!\!\}$.
Because the graph-readout function of graph neural network is injective according to Theorem \ref{thm:1}, we should get $\mathcal{A}(G)\neq \mathcal{A}(H)$, which contradicts our assumption.

For the second part of the conclusion that there exist graph $G$ and $H$ that are decided as isomorphic by 1-WL test but non-isomorphic by subgraph-based graph neural network $\mathcal{A}$, to prove it, we can just find an example that satisfies this. The example shown in Figure \ref{fig:concept}(a) cannot be distinguished by 1-WL graph isomorphism test, but for the subgraph associated with each node, the random walk graph kernel can embed them to different embeddings, by interacting with an appropriate graph filter (e.g., a graph filter with one node).

\subsection{Proof of Lemma 2}
To find at least one feasible $\Phi(H)$, assume the length of non-zero vector $\Phi(G_v)$ is large but finite $\Phi(G_v)=[c_0,c_1,...,c_N]$ with maximum absolute value $c$, then we can encode each value of $\Phi(G_v)$ with the base $2c$ according to their positions in the vector. Specifically, we can let $\Phi(H)=[(2c)^0,(2c)^1,...,(2c)^N]$, and the inner product $K(H,G_v)=\left<\Phi(H),\Phi(G_v)\right>=\sum_{i=0}^{N}c_i(2c)^i$ will be injective with respect to $\Phi(G_v)$.

\section{Connections between KerGNNs and CNNs}
In this section, we discuss the claim in the paper that KerGNNs generalize CNNs into the graph domain from the kernel's point of view. We show in the first subsection that 2-D image convolution in CNNs is equivalent to calculating the appropriate kernel function between patches and filters. Then we show in the second subsection that KerGNNs generalize this aggregation approach by introducing the counterparts of patch, filter and convolution in the graph regime.
\label{sec:rethinking}
\subsection{Rethinking CNNs from the Kernel's Perspective}
Standard CNN models use image convolution to aggregate the local information around each pixel. In this subsection, we show that the image convolution can be viewed as applying kernel functions between input patches and filters in the convolutional layers.

Given a convolutional layer in CNN, the input to the layer is an image $I\subset{\Omega}$, where $\Omega\subset [0, 1]^2$ is a set of pixel coordinates. Typically, $\Omega$ is a two-dimensional grid.
We also define a feature mapping function to map every pixel in the image to a finite vector space, ~$\phi$: $I \rightarrow \mathbb{R}^{d}$, where $d$ is the dimension of the input feature space.
For each pixel $q\in{I}$ with coordinate $(x,y)$, 
we can find a neighborhood patch $\textbf{x}_{q}$ centered at the pixel $q$, with patch size $r\times r$. $\phi(q)$ is the feature map of the pixel in $\mathbb{R}^{d}$, and with some abuse of notation, $\phi(\textbf{x}_{q})$ is defined as the concatenation of feature maps of every pixel in the patch $\textbf{x}_{q}$, i.e., $\phi(\textbf{x}_{q}) = [\phi(q_{j})]_{j=1}^{r\times r}$ for every pixel $q_j\in \textbf{x}_{q}$. Thus, $\phi(\textbf{x}_{q})$ lives in the vector space $\mathbb{R}^{d\times r^2}$. For example, given an RGB image as the input, $\phi(q)$ is in the Euclidean space $\mathbb{R}^3$, and $\phi(\textbf{x}_q)$ is the feature vector in $\mathbb{R}^{3\times {r}^2}$.

\begin{figure*}
  \centering \includegraphics[width=0.85\textwidth]{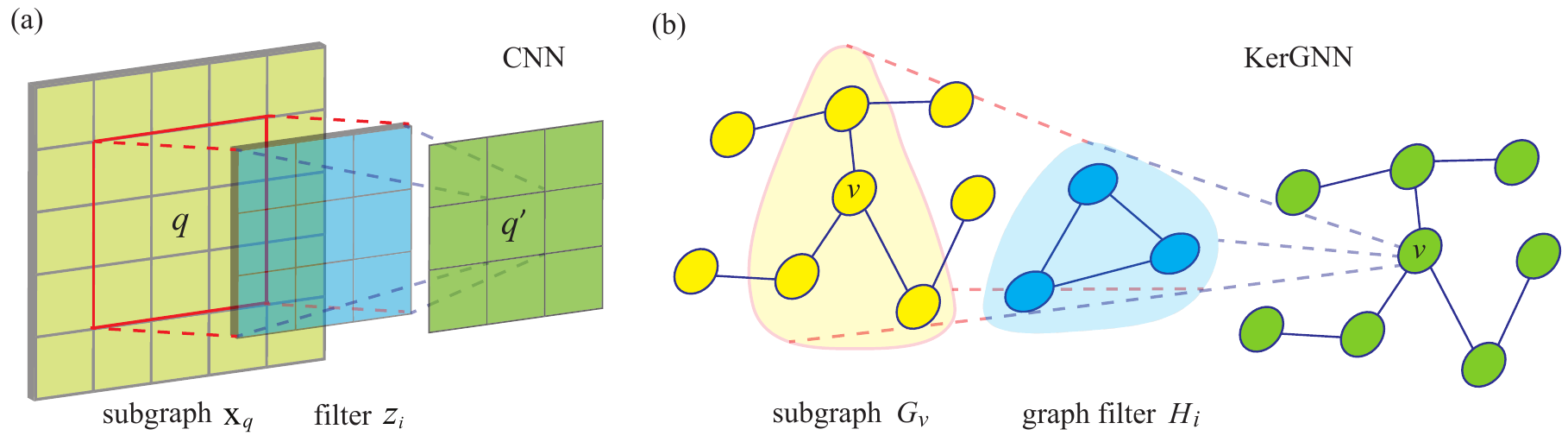}\vspace{-0.1in}
  \caption{\textbf{Comparisons of filters in CNN and KerGNNs.} (a) The yellow grids represent the input image, and the pixels in the red box represent the patch around pixel $q$. The blue grids represent the filter. The position of pixel $q$ in the output grid is denoted by $q'$. (b) The graph denoted in the yellow shadow represents the subgraph of node $v$. The graph with blue shadow represents the graph filter.}\vspace{-0.1in}
  \label{fig:comparison}
\end{figure*}

Next, we represent the output of the layer as a different image $I'\subset \Omega$ with the feature mapping function $\phi'$: $I' \rightarrow \mathbb{R}^{d'}$, where $d'$ is the dimension of the output feature space. As shown in Figure \ref{fig:comparison}(a), $I$ and $I'$ may not have the same size, but every pixel $q'\in{I'}$ corresponds to a pixel $q\in{I}$ with an associated patch $\textbf{x}_{q}$. The goal of the convolutional layer is then to learn this output feature mapping function $\phi'$. Specifically, the convolutional layer adopts $d'$ filters to perform 2-D convolution operation over the image. 
The $i$-th filter performs the dot product over each patch of the image with a fixed stride, and can be parameterized as a vector $z_i\in \mathbb{R}^{{d}\times {r^2}}$. The output feature is obtained by computing the dot product of $z_{i}$ and $\phi(\textbf{x}_q)$ and followed by an element-wise nonlinear function $\sigma$. In other words, it is the $i$-th dimension of the feature representation $\phi'(q')$ of the pixel $q'$ in the new image $I'$. The process can be written as follows:

\begin{align}
        \phi'_{i}(q') & = \sigma(\Phi \ast Z_i)\nonumber\\
                 & =\sigma\left(\sum_{t=1}^{d}\sum_{m,n=1}^{r}{\Phi[t,x-m,y-n] Z_{i}[t,m,n]}\right)\nonumber\\
                 & = \sigma(\langle\phi(\textbf{x}_{q} ) \,, z_{i}\rangle) = \sigma(z_{i}^T \phi(\textbf{x}_{q} )),
    \label{eq:convolution}
\end{align}
where $\ast$ represents the image convolution, $\Phi$ and $Z_{i}$ are tensors with shape $(d\times r\times r)$ reshaped from vectors $\phi(\textbf{x}_{q})$ and $z_{i}$, respectively. We omitted the bias term here for simplicity.

Now, we rethink this process from the kernel perspective. According to the theory of kernel methods, each positive definite kernel function $K$ implicitly defines a RKHS $\mathcal{H}$. Next, we try to make this implicit RKHS to be the output feature space of this convolutional layer, by appropriately designing the associated kernel function. Here, we can define a simple dot-product RBF kernel function between the input feature map vector $\phi(\textbf{x}_q)$ and the filter vector $z_{i}$ as
\begin{equation}
\begin{split}
    K(\phi(\textbf{x}_q),z_{i}) & = \exp(z_{i}^T \phi(\textbf{x}_q)).
\end{split}
\end{equation}
It is worth noting that RKHS $\mathcal{H}$ determined by this RBF kernel is of infinite dimensions, and thus we need to make the assumption here that $\mathcal{H}$ can be approximated by a finite output vector space, with a finite set of basis vectors which are defined as trainable filters $\{{z_{i}: i=1,...,d'\}}$ and $\mathcal{H} = \mathrm{Span}(z_{1},...,z_{d'})$. This is similar to the assumption made by \citep{mairal2014convolutional}. It is noted that the output feature space can also be  approximated by a subspace of $\mathcal{H}$ using Nystr{\"o}m method as in \citet{mairal2016end}.

Then we can derive the output feature map of the pixel $q'$ by projecting the input feature map vector $\phi(\textbf{x}_q)$ into $\mathcal{H}$, and thus the $i$-th dimension of the output feature map $\phi'(q')$ can be computed as the projection of $\phi(\textbf{x}_q)$ onto the $i$-th basis vector $z_{i}$ as
\begin{equation}
\begin{split}
     \phi'_{i}(q') & = \langle\phi(\textbf{x}_{q} ) \,, z_{i}\rangle_{\mathcal{H}} \\
     & = K(\phi,z_{i}) =  \exp(z_{i}^T\phi(\textbf{x}_{q})),\\
\end{split}
\end{equation}
where the second equality holds because of the kernel method. Then we obtain the formula to calculate the output feature mapping function from a perspective that is different from image convolution, and the result is similar to Equation \ref{eq:convolution}. The only difference is in the element-wise nonlinear activation function, and it is worth noticing that the exponential nonlinearity induced by the RBF kernel is very similar to the popular \textit{ReLu} function $\sigma(\cdot)$ used in practice.

Now we come to the conclusion that under suitable assumptions, the standard image convolution in CNN layers can be approximately interpreted as applying kernel functions between the input patches and the trainable filters, and we can use the computed kernel value to update the feature map of the pixel in the output space. This is our major motivation for the design of KerGNN. In KerGNN, we use the computed kernel values to update the feature map of the node in the output graph. This comparison of convolutional layers and KerGNN layers is shown in Figure \ref{fig:comparison}.
\subsection{Generalizing CNNs to KerGNNs}
We explain in this subsection how we derive the update rule shown in Algorithm \ref{alg:kergnn} inspired by CNNs. We consider a single KerGNN layer with an undirected graph $G=(\mathcal{V},\mathcal{E})$ and node feature map of the input graph $G$ can be characterised as $\phi_0: \mathcal{V}\to \mathbb{R}^{d_0}$ as input. 

We rely on the graph filters (as the counterpart of filters) $\{H^{(1)}_{i}, i=1,...,d_1 \}$ to aggregation each node's subgraph (as the counterpart of the patch) and obtain $\phi_1(v)$.
Specifically, we consider using the $\mathcal{R}$-convolution typed kernel function 
which implicitly defines an RKHS $\mathcal{H}_1$. Similar to the analysis of CNN from the kernel perspective (as discussed in the first subsection), we assume that $\mathcal{H}_1$ can be approximated by the finite-dimensional space $\mathbb{R}^{d_1}$ with a set of $d_1$ vectors $\{\Phi(H^{(1)}_{i}),i=1,...,d_1\}$, so that $\mathcal{H}_1=\mathrm{Span}(\Phi(H^{(1)}_{i}),...,\Phi(H^{(1)}_{d_1}))$, 
Then, we calculate $\phi_1(v)$ by projecting subgraph feature map $\Phi_0(G_v)$ into $\mathcal{H}_1$, and the $i$-th dimension of $\phi_1(v)$ can be calculated as the inner product between $\Phi_0(G_v)$ and the basis vector $\Phi(H^{(1)}_{i})$, i.e.,
\begin{equation}
\label{eq:1LayerKernel}
\begin{split}
   \phi_{1,i}(v) = \langle \Phi_0(G_{v}),\Phi(H^{(1)}_{i})\rangle_{\mathcal{H}_1}
   \approx K(G_v,H^{(1)}_i),
\end{split}
\end{equation}
where we adopt random walk kernel as $K$. Similar to our conclusions for CNNs, Equation \ref{eq:1LayerKernel} indicates that we can use the kernel values computed by the subgraph of a node and graph filters as the output feature map of the node. After calculating the kernel value of the subgraph $G_v$ with respect to every graph filter $\{H^{(1)}_{i},i=1,...,d_1\}$, we obtain every dimension of the feature map of node $v$ in $\mathcal{H}_1$, which is the output of the KerGNN layer.

We compare CNNs and KerGNNs in Figure \ref{fig:comparison}, and summarize the similarities in the following:

\textbf{Sliding over Inputs.} In CNN, the filter is systematically applied with each filter-sized patch of the input image, from left to right and top to bottom, with a specified stride. In KerGNNs, we sample a subgraph $G_{v}=(\mathcal{V}_v,\mathcal{E}_v)$ consisting of $v$ and its $j$-hop neighbors, and the graph filter is applied to ${G_{v}}$ for $\forall v\in \mathcal{V}$ (the adjacency of $G$ is preserved in the subgraph). It should be noted that the operations defined below do not require the subgraph and graph filter have the same number of nodes or topology.

\textbf{Shared Parameters.} In CNN, all the patches share the same filter for convolution to reduce the number of parameters. In KerGNNs, all the subgraphs $G_{v}$ for $\forall v\in \mathcal{V}$ also share the same graph filter, and thus the parameters of graph filters will not scale up as the input graph becomes larger.

\textbf{Local Aggregation.} In CNN, the patch consists of a central pixel and neighboring pixels, and the feature map of the patch $\phi (\textbf{x}_{p})$ contains all the neighbors' information. The filters aggregate the patch and assign the corresponding kernel value to the output feature map of the central pixel. In KerGNNs, we use the graph filters to aggregate the feature map of the subgraph $\Phi (G_{v})$, and let the kernel value be the output feature map of the central node.

\section{Connections between KerGNNs and MPNNs}
We show in this section that KerGNNs generalize the standard MPNNs. From the point view of KerGNNs, MPNNs deploy a simple graph filter with one node, and an appropriate kernel function can be chosen with KerGNN framework, such that KerGNNs can iteratively update nodes' representations using neighborhood aggregation like in MPNNs. For example, the vertex update rule of Graph Convolutional Network (GCN) \citep{kipf2016semi} can be written as
\begin{equation}
\label{eq:gcn}
\begin{split}
   \phi_{l}(u) = \sigma\left(W\sum_{v\in \mathcal{N}(u)\cup \{ u\}}\frac{\phi_{l-1}(v)}{\sqrt{|\mathcal{N}(u)|\cdot|\mathcal{N}(v)|}}\right),
\end{split}
\end{equation}
where $\mathcal{N}(u)$ represents all the neighboring nodes of $u$, $\sigma$ is the element-wise nonlinear function, $\phi_{l-1}(v)\in \mathbb{R}^{d_{l-1}}$ and $\phi_{l}(v)\in \mathbb{R}^{d_{l}}$. To interpret this updating rule in the KerGNN framework, we can define the subgraph $G_u$ containing nodes $\mathcal{N}(u)\cup \{ u\}$. We also define $d_l$ graph filters $\{H^{(l)}_{i},i=1,...,d_l \}$, and each graph filter is defined as $H^{(l)}_{i}=(\{h_i \}, \{ \})$. The attribute of the node $h_i$ is parameterized by $W^{(l)}_{i}\in \mathbb{R}^{1\times{d_{l-1}}}$. Then we define an $\mathcal{R}$-convolution graph kernel as
\begin{equation}
\begin{split}
  K(G_u,H^{(l)}_{i}) &= \sum_{v\in \mathcal{N}(u)\cup \{ u\}}\sum_{v'\in  \{h_i\}} k_{\mathrm{base}}(v,v')\\
  &=\sum_{v\in \mathcal{N}(u)\cup \{ u\}}k_{\mathrm{base}}(v,h_i),
\end{split}
\end{equation}
and the node-wise kernel $k_{\mathrm{base}}$ is defined as
\begin{equation}
\begin{split}
  k_{\mathrm{base}}(v,h_i) = \sigma\left(\frac{{W^{(l)}_{i}}^T\phi_{l-1}(v)}{\sqrt{|\mathcal{N}(u)|\cdot|\mathcal{N}(v)|}}\right).
\end{split}
\end{equation}
Using the KerGNN framework, the $i$-th dimension of the output feature map $\phi_{l,i}(u)$ can be written as the kernel function between $G_v$ and $H^{(l)}_{i}$:
\begin{align}
  \!\!\!\!\phi_{l,i}(u) &= K(G_u,H^{(l)}_{i})\nonumber\\
  &=\sum_{v\in \mathcal{N}(u)\cup \{ u\}}\sigma\left(\frac{{W^{(l)}_{i}}^T\phi_{l-1}(v)}{\sqrt{|\mathcal{N}(u)|\cdot|\mathcal{N}(v)|}}\right),\nonumber\\
  &=\sigma\!\left({W^{(l)}_{i}}^T\!\!\!\!\!\sum_{v\in \mathcal{N}(u)\cup \{ u\}}\!\frac{\phi_{l-1}(v)}{\sqrt{|\mathcal{N}(u)|\cdot|\mathcal{N}(v)|}}\right)\!,
\end{align}
which is equivalent to Equation \ref{eq:gcn}. Therefore, the message aggregation in most MPNNs can be treated as using one-node graph filters in KerGNNs, and our proposed method generalizes MPNNs by deploying more complex graph filters with multiple nodes and learnable adjacency matrix.

\section{Model Implementation Details}
To construct the subgraph for a node, we first calculate all the $j$-hop neighbors of the node, and extract the subgraph determined by the node and all its neighbors. In implementation, to be compatible with matrix multiplication, we set a maximum size of subgraphs. Any subgraph exceeding this limit is truncated and preserves nearer neighbors. The adjacency matrix of a subgraph that does not reach this limit will be padded with zeros. Therefore, the adjacency matrices of all the subgraphs have the same size.

For the first layer of the KerGNN model, we optionally add an additional linear mapping to transform node attributes of the input graph to a specified dimension (the dimension of node attributes of graph filter in the first layer), such that we can calculate the graph kernel between input graphs and graph filters at the first layer.

For simplicity, we define all the graph filters at the same layer to have the same number of nodes. Besides, for comparison, we only use one type of graph kernels within one model, although different graph filters can interact with same subgraphs with different types of graph kernels and thus different graph kernels can be mixed within one model.

\section{Experiment Details and More Results}
\subsection{Graph Classification Task}
\subsubsection{Hyper-parameter Search.}
We conduct the experiment using Intel(R) Core(TM) i7-7700 CPU @ 3.60GHz CPU with NVIDIA GPU (GeForce RTX 2070). We use grid search to select the hyper-parameters for each model during cross-validation, and the hyper-parameter search range is shown in Table \ref{table-hyperparam}.
\begin{table}[ht]
\caption{\textbf{Hyper-parameter search range}.}
\label{table-hyperparam}
\begin{center}
\begin{tabular}{lc}
\toprule
hidden dimension of the first layer & [8; 16; 32; 64]\\
number of graph filter & [16; 32; 64; 128]  \\
number of nodes of graph filter & [2; 4; 6; 8; 10; 12; 14; 16; 18; 20]\\
maximum number of nodes for subgraph& [5; 10; 15; 20; 25; 30]\\
$j$-hop neighborhood & [1; 2; 3]\\
maximum step for random walk &[1; 2; 3; 4; 5]\\
hidden dimension of linear layer & [8; 16; 32; 48; 64]\\
dropout rate & [0.2; 0.4; 0.6; 0.8]\\
\bottomrule
\end{tabular}
\end{center}
\vskip -0.1in
\end{table}

\subsubsection{Details of Datasets.}
We use 5 bioinformatics datasets: MUTAG is a dataset of 188 mutagenic aromatic and heteroaromatic nitro compounds with 7 discrete labels.
PROTEINS dataset uses secondary structure elements (SSEs) as nodes and two nodes are connected if they are neighbors in the amino-acid
sequence or in 3D space. It has 3 discrete labels, representing helix, sheet, or turn. 
NCI1 is a dataset made publicly available by the National Cancer Institute (NCI) and is a subset of balanced datasets of chemical compounds screened for ability to suppress or inhibit the growth of a panel of human tumor cell lines.
DD is a dataset of 1178 protein X-Ray structures. Two nodes in a protein are connected by an edge if they are less than 6 Angstroms apart. The prediction task is to classify the protein structures into enzymes and non-enzymes.

We use 4 social datasets:
IMDB-BINARY and IMDB-MULTI are movie collaboration datasets. Each graph corresponds to an ego-network for each actor/actress, where nodes correspond to actors/actresses and an edge is drawn between two actors/actresses if they appear in the same movie. Each graph is derived from a pre-specified genre of movies, and the task is to classify the genre that the graph is derived from. IMDB-BINARY considers two genres: Action and Romance. IMDB-MULTI considers three classes: Comedy, Romance, and Sci-Fi.
Each graph of REDDIT-BINARY dataset corresponds to an online discussion thread and nodes correspond to users. Two nodes are connected if at least one of them responded to another’s comment. The task is to classify each graph to a community or a subreddit it belongs to. 
COLLAB is a scientific collaboration dataset, derived from 3 public collaboration datasets, namely, High Energy Physics, Condensed Matter Physics, and Astro Physics. Each graph corresponds to an ego-network of different researchers from each field. The task is to classify each graph to a field the corresponding researcher belongs to.

\begin{figure*}[h]
  \centering \includegraphics[width=1.0\textwidth]{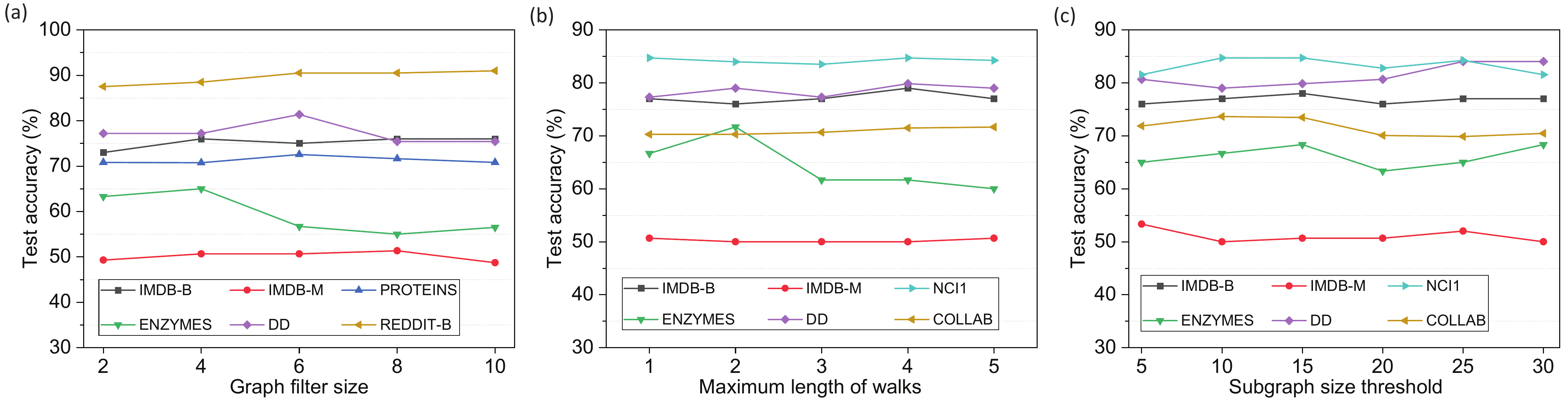}
  \caption{\textbf{Test accuracy w.r.t. model parameters}. The results are obtained from experiments on 1 fold of datasets, and for all the three graphs only the model parameter on the x-axis is changed with remaining model parameters fixed.}
  \label{sfig:paramstudy}
\end{figure*}
\subsubsection{Model Parameters.}
We study how the test accuracy is influenced by the number of nodes in the graph filters, which determines the size and complexity of the graph filters. As shown in Figure \ref{sfig:paramstudy}(a), the optimal size of the graph filter is different for different datasets, depending on the local structures of different types of graphs, e.g., the star patterns in graphs of REDDIT-B and the ring and chain patterns in graphs of NCI1. We also study the influence of the maximum length of random walks as shown in Figure \ref{sfig:paramstudy}(b), it can be seen that longer walks generally benefit the classification results except for ENZYMES dataset where the model performs the best with walk of length 2. In the code implementation, we specify the maximum number of nodes that the subgraph can contain, which implicitly control the size of the subgraphs, and thus we also study the influence of this threshold of node numbers in Figure \ref{sfig:paramstudy}(c). DD and ENZYMES achieve higher accuracy with larger subgraphs, because larger subgraph contains more fruitful neighborhood topology information. The remaining datasets are not influenced too much, because we fix the size of graph filters, and the model performance degrades when the subgraph size and graph filter size mismatch, for example, small graph filters cannot handle larger subgraphs.

\subsubsection{Model Running Time.}
We compare the running time of the proposed model with several GNN benchmarks, as shown in Table \ref{table-time}. We measure the one-epoch running time averaged over 50 epochs, using the same GPU card. All the models are set to have hidden dimension 32 and 1 MLP layer. For KerGNNs, the number of nodes in the graph filter is set to 6 and the maximum subgraph size is set to 10.
We observe that KerGNN gives much better running time compared with high-order GNN benchmarks, and achieves similar performance compared with conventional GNN models with 1-WL constraints.

\begin{table}[ht]
\caption{\textbf{The measured one-epoch running time (s) of different GNN models}.}
\label{table-time}
\begin{center}
\begin{small}
\begin{sc}
\resizebox{\textwidth}{!}{
\begin{tabular}{lcccccccc}
\toprule
 & DD & NCI1 & PROTEINS & ENZYMES & IMDB-B & IMDB-M & REDDIT-B & COLLAB \\
\midrule
DGCNN      & 0.271$\pm$0.005 & 0.425$\pm$0.009 & 0.134$\pm$0.006 & 0.069$\pm$0.002 & 0.113$\pm$0.007 & 0.161$\pm$0.008 & 0.411$\pm$0.010 & 0.824$\pm$0.053 \\
DiffPool   & 2.887$\pm$0.029 & 4.232$\pm$0.035 & 1.125$\pm$0.006 & 0.131$\pm$0.007 & 0.210$\pm$0.031 & 0.295$\pm$0.013 & 9.580$\pm$0.233 & 2.296$\pm$0.088 \\
ECC        & 110.14$\pm$ 4.31 & 1.179$\pm$0.012 & 0.550$\pm$0.012 & 0.233$\pm$0.005 & 0.254$\pm$0.007 & 0.304$\pm$0.050 &  OOR & OOR \\
GIN        & 0.344$\pm$0.003 & 0.443$\pm$0.007 & 0.133$\pm$0.006 & 0.072$\pm$0.004 & 0.118$\pm$0.032 & 0.160$\pm$0.008 & 0.740$\pm$0.053 & 0.809$\pm$0.044 \\
GraphSAGE  & 0.141$\pm$0.002 & 0.331$\pm$0.014 & 0.088$\pm$0.003 & 0.047$\pm$0.002 & 0.081$\pm$0.004 & 0.122$\pm$0.009 & 0.198$\pm$0.011& 0.589$\pm$0.026 \\
\midrule
1-2-3 GNN  & OOR & 1.228$\pm$0.075 & 1.056$\pm$0.058 & OOR & 1.301$\pm$0.071 & 1.754$\pm$0.075 & 0.608$\pm$0.025 \footnote{This result is obtained from 1-GNN model.}  & OOR\\
Powerful GNN  & OOR & 19.764$\pm$0.270 & 4.24$\pm$0.23 & 1.997$\pm$0.150 & 2.701$\pm$0.084 & 3.417$\pm$0.092 & OOR  & 17.578$\pm$0.621\\
\midrule
KerGNN-1  & 0.732$\pm$0.041 & 0.445$\pm$0.033 & 0.183$\pm$0.034 & 0.370$\pm$0.030 & 0.092$\pm$0.039 & 0.086$\pm$0.026 & 0.865$\pm$0.051  & 1.336$\pm$0.031\\
KerGNN-2  & 1.486$\pm$0.078 & 0.688$\pm$0.038 & 0.373$\pm$0.043 & 0.085$\pm$0.019 & 0.148$\pm$0.044 & 0.225$\pm$0.036 & 1.677$\pm$0.051  & 2.634$\pm$0.047\\
KerGNN-3  & 2.748$\pm$0.078 & 0.876$\pm$0.044 & 0.553$\pm$0.026 & 0.126$\pm$0.024 & 0.196$\pm$0.019 & 0.343$\pm$0.030 & 2.551$\pm$0.065  & 3.944$\pm$0.046\\
KerGNN-DRW  & 1.782$\pm$0.051 & 0.753$\pm$0.034 & 0.405$\pm$0.045 & 0.090$\pm$0.025 & 0.155$\pm$0.036 & 0.240$\pm$0.036 & 2.078$\pm$0.060  & 2.648$\pm$0.039\\
\bottomrule
\end{tabular}}
\end{sc}
\end{small}
\end{center}
\end{table}

\subsection{Node Classification Task}
We further evaluate the proposed KerGNN model for node classification task. we use 4 datasets: Cora, Citeseer, Pubmed \cite{sen2008collective}, Chameleon \cite{rozemberczki2021multi}. For each dataset, we randomly split nodes of each class into 60\%, 20\%, and 20\% for training, validation and testing, and report the mean accuracy of all models on the test sets over 10 random splits. We compare our model with several popular GNNs including GCN, GAT \cite{velivckovic2017graph}, GEOM-GCN \cite{pei2020geom}, APPNP \cite{klicpera2018predict}, JKNet \cite{xu2018representation} and GCNII \cite{chen2020simple}. As shown in Table\ref{table-nodecla}, KerGNNs achieve similar or better results compared with SOTA baselines.

\begin{table}[ht]
\caption{\textbf{Node classification accuracy}.}
\label{table-nodecla}
\begin{center}
\begin{sc}
\begin{tabular}{lcccccccc}
\toprule
 & Cora & Citeseer & Pubmed & Chameleon\\
\midrule
GCN   & 85.77 & 73.68 & 88.13 &28.18\\
GAT   & 86.37 & 74.32 & 87.62 &42.93\\
Geom-GCN & 85.27 & \textbf{77.99} & \textbf{90.05}&60.90 \\
APPNP   & 87.87 & 76.53 & 89.40&54.30\\
JKNet  & 87.46 & 76.83 & 89.18&62.08\\
GCNII & \textbf{88.49} & 77.08 & 89.57 &60.61\\
\midrule
KerGNN  & 87.96 & 76.61 & 89.53 & \textbf{62.28}\\
\bottomrule
\end{tabular}
\end{sc}
\end{center}
\end{table}

\section{More Visualizations}
\begin{figure*}[h]
  \centering \includegraphics[width=0.8\textwidth]{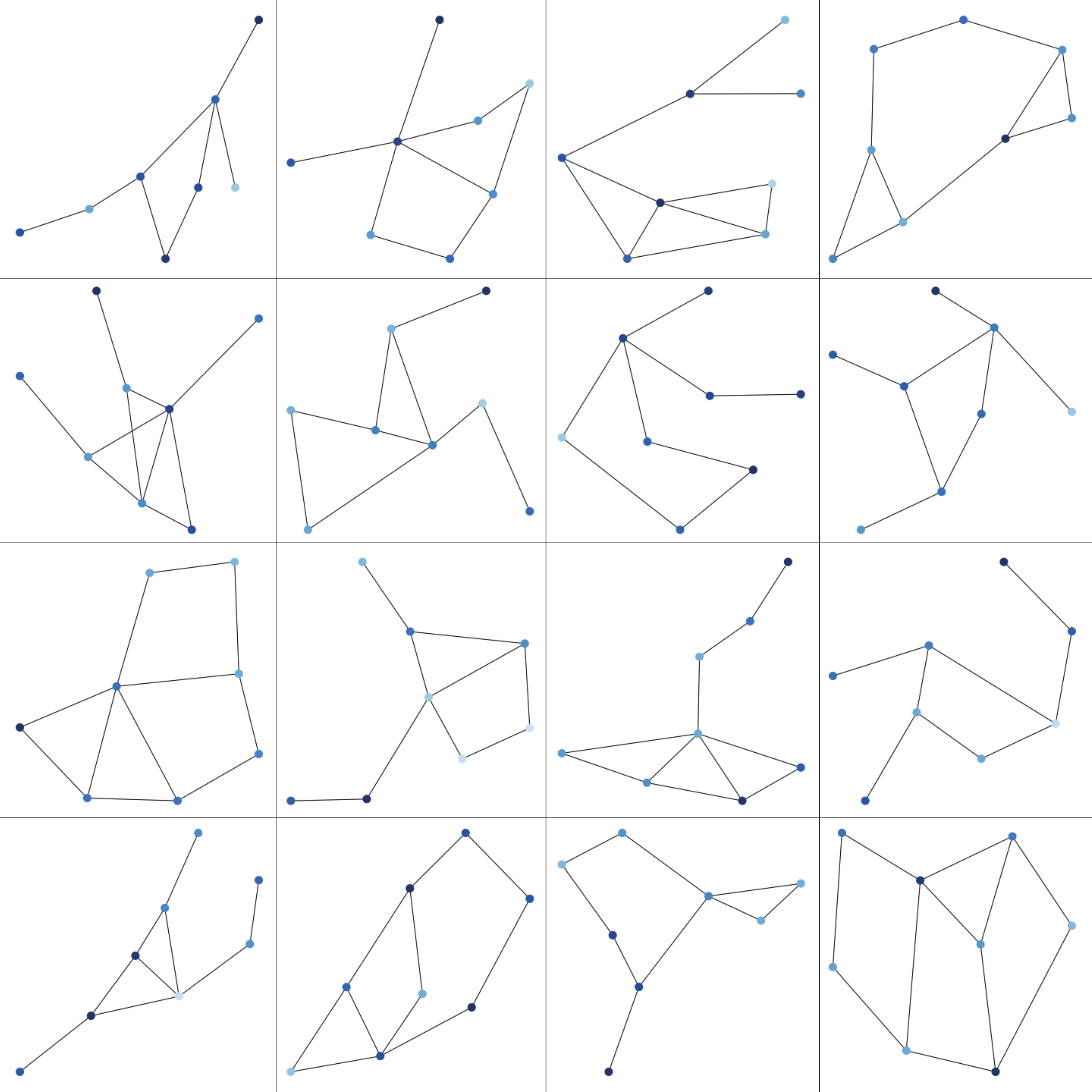}
  \caption{\textbf{Graph filters of MUTAG}}\vspace{-0.1in}
  \label{fig:mutag}
\end{figure*}

\begin{figure*}[tb]
  \centering \includegraphics[width=0.8\textwidth]{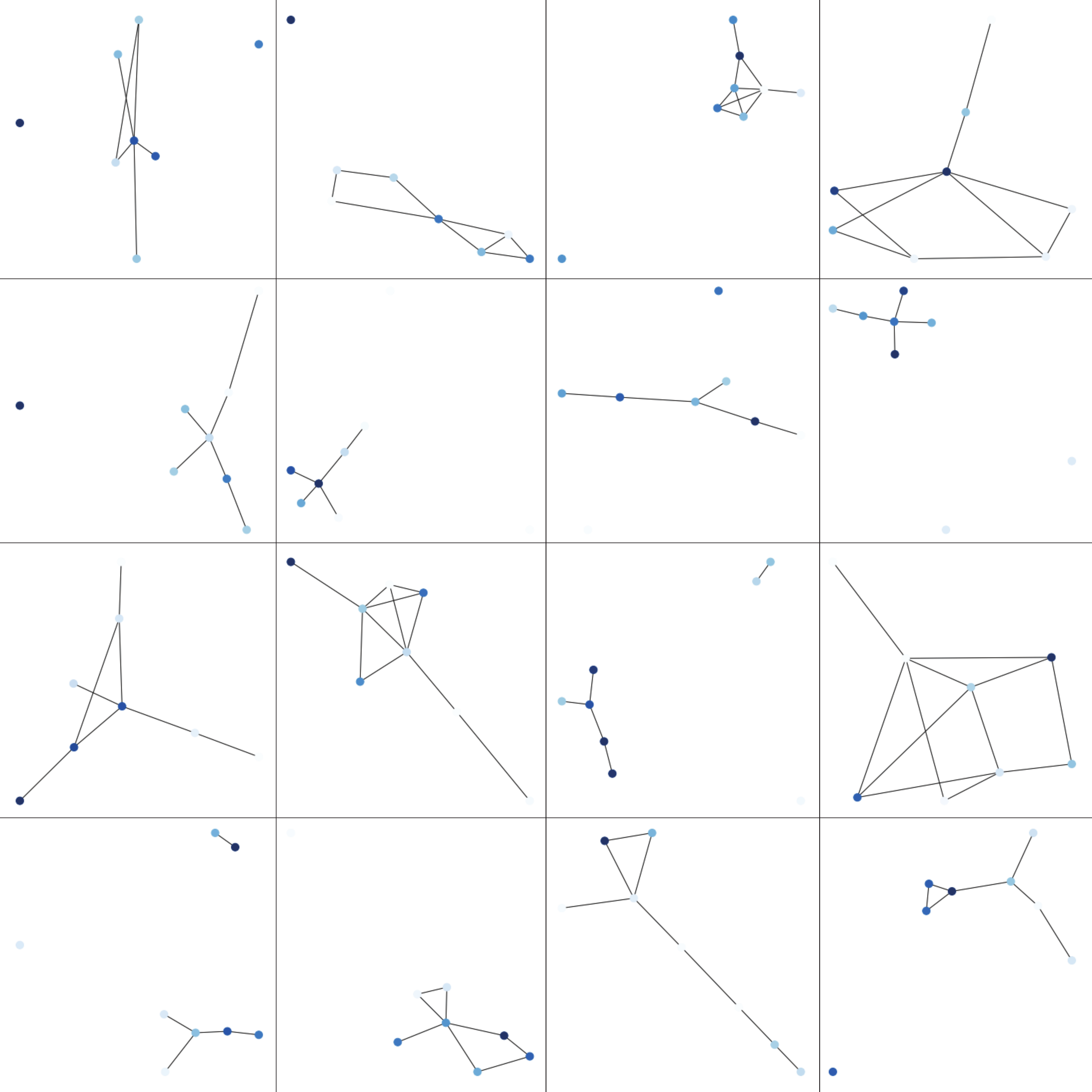}
  \caption{\textbf{Graph filters of REDDIT-BINARY}}\vspace{-0.1in}
  \label{fig:reddit}
\end{figure*}

In this section, we show the graph filters trained for MUTAG dataset and REDDIT dataset in Figures \ref{fig:mutag} and \ref{fig:reddit}. We can see that the trained graph filters have different patterns for the two datasets, and each type of patterns reveals the characteristics of corresponding dataset. Specifically, graph filters for MUTAG tend to have ring and circular patterns, while graph filters for REDDIT tend to have star patterns.

\end{document}